\documentclass{article} % For LaTeX2e
\usepackage{iclr2027_conference,times}

\usepackage{amsmath,amsfonts,bm}

\def\eqref#1{equation~\ref{#1}}
\def\1{\bm{1}}

\DeclareMathAlphabet{\mathsfit}{\encodingdefault}{\sfdefault}{m}{sl}
\SetMathAlphabet{\mathsfit}{bold}{\encodingdefault}{\sfdefault}{bx}{n}

\usepackage{hyperref}
\usepackage{url}

\usepackage{graphicx}
\usepackage{xspace}
\usepackage{pifont}
\usepackage{algorithm}
\usepackage{algpseudocode}
\usepackage{xcolor}
\usepackage{booktabs}
\usepackage{multirow}
\usepackage{graphicx}
\usepackage{arydshln}
\usepackage{graphicx}
\usepackage{subcaption}
\usepackage[export]{adjustbox}

\usepackage{float}
\usepackage{wrapfig}
\usepackage{algorithm}
\usepackage{algpseudocode}

\usepackage[most]{tcolorbox}
\usepackage{listings}
\usepackage{xcolor}
\usepackage{tabularx}
\usepackage{enumitem}
\usepackage{booktabs}

\usepackage[most]{tcolorbox}
\usepackage{listings}
\usepackage{xcolor}
\usepackage{enumitem}

\definecolor{PromptBg}{HTML}{F7F8FA}
\definecolor{PromptFrame}{HTML}{A9B4C2}
\definecolor{PromptTitle}{HTML}{D8DEE6}

\lstdefinestyle{jsonstyle}{
  basicstyle=\ttfamily\scriptsize,
  breaklines=true,
  breakatwhitespace=false,
  columns=fullflexible,
  keepspaces=true,
  showstringspaces=false,
  frame=none,
  tabsize=2
}

\newtcolorbox{promptbox}[1]{
  enhanced,
  title={#1},
  colback=PromptBg,
  colframe=PromptFrame,
  colbacktitle=PromptTitle,
  coltitle=black,
  fonttitle=\bfseries\small,
  boxrule=0.5pt,
  arc=2pt,
  left=6pt,
  right=6pt,
  top=5pt,
  bottom=5pt,
  before skip=6pt,
  after skip=6pt,
  fontupper=\small
}

\definecolor{SkillBg}{HTML}{F4FAF7}
\definecolor{SkillFrame}{HTML}{7BAE94}
\definecolor{SkillTitle}{HTML}{DCEFE5}

\newtcolorbox{skillbox}[1]{
  enhanced,
  title={#1},
  colback=SkillBg,
  colframe=SkillFrame,
  colbacktitle=SkillTitle,
  coltitle=black,
  fonttitle=\bfseries\small,
  boxrule=0.5pt,
  arc=2pt,
  left=6pt,
  right=6pt,
  top=5pt,
  bottom=5pt,
  before skip=6pt,
  after skip=6pt,
  fontupper=\small
}

\newcommand{\promptfield}[1]{\vspace{0.35em}\noindent\textbf{#1}\par\vspace{0.15em}}

\definecolor{commentteal}{RGB}{30,120,120}
\algrenewcommand\algorithmiccomment[1]{\hfill{\scriptsize\textcolor{commentteal}{$\triangleright$ #1}}}

\newcommand{\Name}{\texttt{CODESKILL}\xspace}

\title{\textsc{CODESKILL}: Learning Self-Evolving Skills for Coding Agents}

\iclrfinalcopy
\author{%
\parbox{\dimexpr\textwidth-2\tabcolsep\relax}{%
\centering
Yanzhou Li$^1$, Yiran Zhang$^1$, Xiaoyu Zhang$^1$,
Xiaoxia Liu$^2$, and Yang Liu$^1$\\
\normalfont
$^1$Nanyang Technological University,
$^2$Zhejiang University\\
\texttt{\{yanzhou001, yiran002, xiaoyu.zhang, yangliu\}@ntu.edu.sg}\\
\texttt{liuxiaoxia@zju.edu.cn}
}}

\begin{document}

\maketitle
\lhead{Preprint}
\begin{abstract}
Coding agents produce rich trajectories while solving software-engineering tasks. To enable agent self-evolution, these trajectories can be distilled into reusable procedural skills that compactly encode experience to guide future behavior.
However, existing skill construction and maintenance methods often rely on fixed prompts and heuristic update rules, leaving it unclear how knowledge should be selected, abstracted, and maintained to best serve downstream agents. We propose \Name, an LLM-based framework that reformulates skill extraction and skill-bank maintenance as a learnable management policy. \Name extracts multi-granularity procedural skills from coding-agent trajectories, evolves skills with new experience, and maintains a compact skill bank for future task solving. We train \Name with reinforcement learning, using a hybrid reward that combines dense rubric-based skill-quality feedback with sparse verifiable execution feedback from the frozen downstream agent. Experiments on EnvBench, SWE-Bench Verified, and Terminal-Bench 2 show that \Name improves average pass rate by 11.03 over the no-skill baseline and by 5.10 over the strongest prompt-based or memory baseline, while maintaining a compact skill bank.
\end{abstract}
\section{Introduction}
% Large language model (LLM) agents have shown strong capabilities in complex interactive tasks, including coding agents that resolve realistic software issues through repository inspection, test execution, code editing, and failure recovery. These long-horizon interactions produce rich trajectories with reusable experience beyond individual tasks. To reuse such experience, recent agent systems explore memory mechanisms that store prior episodes, retrieve relevant cases, or summarize past trajectories \citep{chen2025sweexp,tang2025agentkb,zhu2026swecontextbench,shen2026subtaskmemory}. Beyond raw episodic memory, higher-level procedural knowledge such as \emph{skills} provides a more compact and reusable way to specify when a recurring situation arises and what actions should be taken.
Large language model (LLM) agents have shown strong capabilities in complex interactive tasks, such as coding agents that resolve realistic software issues. These long-horizon interactions produce rich trajectories with reusable experience beyond individual tasks. To reuse such experience, recent agent systems explore memory mechanisms that store prior episodes, retrieve relevant cases, or summarize past trajectories~\citep{chen2025sweexp,tang2025agentkb,zhu2026swecontextbench,shen2026subtaskmemory}. Beyond raw episodic memory, higher-level procedural knowledge such as \emph{skills} provides a more compact and reusable way to specify when a recurring situation arises and what actions should be taken. Skill-based agents and skill libraries have been shown effective in long-horizon and tool-use settings, making skills an important form of reusable agent knowledge for improving future behavior~\citep{wang2023voyager,ling2026agentskills,liang2026skillnet}.

Inspired by the effectiveness of skills, recent work has begun to automate the distillation of skills and other reusable knowledge from past trajectories to support self-evolving agents. Across general agents and coding agents, these methods use LLMs to analyze prior successes and failures, extract reusable lessons, workflows, or reasoning patterns, and update memory over time for future task solving~\citep{ni2026trace2skill,xia2026skillrl,yang2026autoskill,chen2025sweexp,ouyang2026reasoningbank}.

Despite this progress, existing approaches still rely heavily on fixed prompts and heuristic criteria~\citep{xia2026skillrl,yang2026autoskill,shen2026subtaskmemory}, which predefine what to extract and how to update memory rather than adapting these decisions to the downstream agent. This limitation is especially salient for software-engineering (SWE) agents, where long-horizon, tool-driven trajectories contain dense and heterogeneous feedback, making it difficult to distinguish reusable procedural knowledge from task-specific or accidental details. As a result, it remains unclear what evidence should be extracted, how abstract a skill should be, and how the skill bank should be revised so that the resulting skills are both reusable across tasks and actionable for concrete agent decisions.

% Recent work has therefore begun to automate the distillation of skill from past trajectories. These methods use LLMs to analyze prior successes and failures, extract reusable lessons, workflows, or reasoning patterns, and update memory over time to support self-evolving agents \citep{ni2026trace2skill,xia2026skillrl,yang2026autoskill,chen2025sweexp,ouyang2026reasoningbank}.Although these artifacts are named differently in settings, they share the goal of transforming raw trajectories into reusable knowledge to improve future task solving. 

% Despite this progress, existing approaches often rely on fixed prompts and heuristic criteria to extract skills and maintain skill banks. As a result, the key management choices are usually embedded in human-designed procedures, and have not been systematically studied as decisions that shape the utility of the resulting skills. Effective skill management requires selecting useful evidence, choosing the right granularity, and deciding when to revise, merge, or discard skills. These choices are critical because skills that are too broad provide little task-level guidance, while overly specific or redundant skills may fail to transfer and waste context. Motivated by these challenges, we reformulate skill extraction and maintenance as a learnable management policy, with the goal of finding skill generation and update strategies that best suit the downstream agent.

To this end, we propose \Name, an LLM-based framework optimized to generate reusable skills from coding-agent trajectories and maintain a skill bank for augmenting future software-engineering tasks. Given past trajectories and an existing skill bank, \Name extracts reusable procedural knowledge as skills, evolves existing skills based on new or failed experience, and maintains the skill bank by adding useful candidates, merging redundant ones, or dropping unhelpful skills. To train \Name, we first warm-start it with supervised data constructed from coding-agent trajectories and teacher-generated skill operations, and then optimize it with Group Relative Policy Optimization (GRPO) using feedback from a frozen downstream coding agent~\citep{deepseekmath2024grpo}. The reward combines verifiable task feedback on whether the skill helps the downstream agent solve the task with rubric-based judgments of skill quality and behavior-skill alignment~\citep{llmasjudge2023}. This enables \Name to learn a management policy for extracting, refining, and updating skills that remain reusable and actionable.

We validate \Name by instantiating it with Qwen3.5-4B and evaluating it on EnvBench, SWE-Bench Verified, and Terminal-Bench 2. Across these benchmarks, \Name improves average pass rate by 11.03 over the no-skill baseline and by 5.10 over the strongest prompt-based skill extraction or memory-construction baseline, corresponding to relative gains of about 37.6\% and 14.5\%, respectively. The gains hold across both in-domain and out-of-distribution software-engineering tasks.

Our contributions are summarized as follows.
\begin{itemize}
    \item We propose \Name, an LLM-based framework that analyzes coding-agent trajectories across diverse tasks, extracts reusable skills, and maintains an evolving skill bank for software-engineering tasks.
    \item We reformulate skill management as a learnable management policy, and train \Name with reinforcement learning. Our reward combines sparse verifiable feedback with dense rubric-based judgments, balancing executable outcomes with informative supervision when task success is sparse.
    \item We validate \Name on EnvBench, SWE-Bench Verified, and Terminal-Bench 2, showing that it achieves state-of-the-art performance among memory-based methods.
\end{itemize}

\section{Related Work}

\subsection{Self-evolving Long-term Memory}

Long-term memory enables LLM agents to retain experience beyond a single interaction and improve over time. Early memory-augmented agents store reflections, past experiences, or persistent user information for later retrieval \citep{shinn2023reflexion,zhao2023expel,chhikara2025mem0,zhou2025memento}. Recent self-evolving agents further learn utility estimates for episodic memories, distill trajectories into reusable principles, or construct reasoning and workflow memories from successes and failures \citep{zhang2026memrl,wu2025evolver,ouyang2026reasoningbank,wang2025awm}. Procedural memory summarizes agent behaviors into reusable action knowledge \citep{fang2025memp}. 
% UMEM trains a model to extract and manage general memories with semantic-neighborhood utility \citep{ye2026umem}, but its key-value memories lack the multi-granularity procedural operations needed to maintain verifiably grounded skills for long-horizon SWE tasks. 
Beyond general agent settings, several works augment software-engineering agents with memories distilled from past repair trajectories or shared across agent frameworks \citep{chen2025sweexp,tang2025agentkb}. Another line studies how coding agents can retrieve and reuse related issue contexts or subtask-level histories \citep{zhu2026swecontextbench,shen2026subtaskmemory}.
 These works show that coding agents can benefit from past experience, but they mainly retrieve or summarize prior cases. \Name instead learns to extract, evolve, and maintain procedural memories using downstream feedback.

\subsection{Agent Skills \& Automatic Skill Management}

Agent skills have emerged as a procedural form of memory that augments agents with reusable workflows, instructions, scripts, or tool-use knowledge. Unlike episodic memories that mainly recall past interactions, skills are designed to provide actionable guidance when similar situations arise \citep{wang2023voyager,ling2026agentskills,liang2026skillnet}. Recent work has therefore studied automatic skill construction from trajectories and interaction histories, where agents distill reusable lessons, workflows, or failure patterns into skill banks that can be reused and updated over time \citep{ni2026trace2skill,yang2026autoskill,zheng2025skillweaver,alzubi2026evoskill,zhang2026coevoskills,ma2026skillclaw,zhang2026skillflow}.
Several methods further combine skills with reinforcement learning or agent self-improvement loops \citep{xia2026skillrl,wang2025sage,tu2026d2skill,li2026arise}. These works typically improve the task-solving policy with skills, train agents to use skill libraries, or co-evolve skills with the solver. In contrast, \Name focuses on skill management itself, learning how to generate and maintain skills that are adapted to a frozen downstream coding agent. Recent studies on software-engineering skills also show that static skill injection does not always improve coding agents, and that skill usefulness depends strongly on task fit and context \citep{han2026sweskillsbench}. This further motivates adaptive skill management rather than relying on a fixed skill management strategy.

\section{\Name}
\subsection{Problem Formalization}

\begin{figure}[t]
    \centering
    \includegraphics[width=1.0\columnwidth]{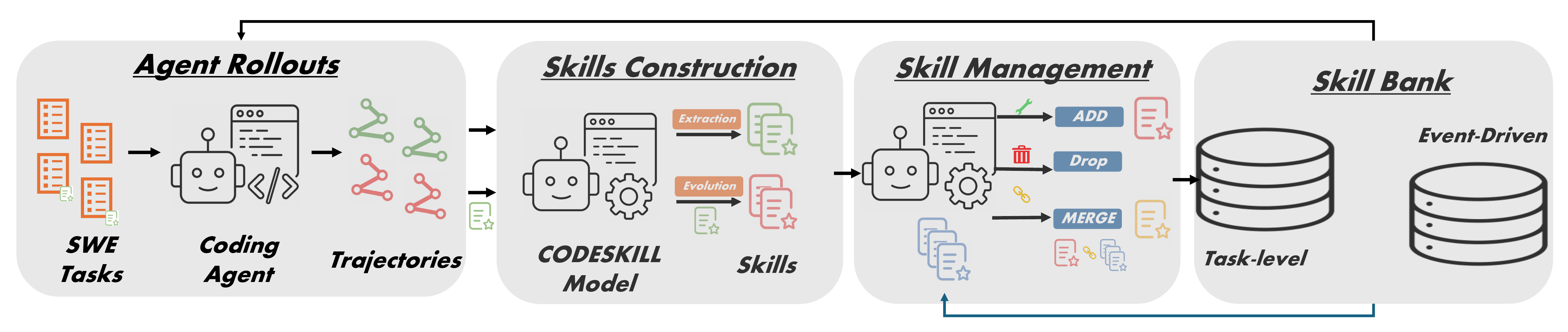}
    \caption{Overview of the \Name pipeline.}
    \label{fig:pipeline}
\end{figure}

We consider a frozen downstream coding policy $\pi$ that solves a software-engineering task $x \in \mathcal{X}$ by interacting with a repository or terminal environment. A rollout produces a trajectory $\tau = (o_1,a_1,\ldots,o_T,a_T,y)$, where $o_t$ is an observation, $a_t$ is an action such as a shell command or file edit, and $y$ is the task outcome. Our goal is to learn from such trajectories without updating $\pi$.

\Name learns a policy $M_\theta$ for managing a skill bank $\mathcal{B}=\{s_i\}_{i=1}^{N}$, where each skill $s_i$ contains reusable procedural instructions that provide actionable guidance. Given trajectory evidence $\tau$ and relevant skill-bank context $\mathcal{C} \subseteq \mathcal{B}$, $M_\theta$ outputs an operation $u=(a,z)$, where $a \in \mathcal{A}$ denotes the operation type, such as generation, evolution, or maintenance, and $z$ denotes the operation content, such as a generated skill or a maintenance decision. We provide the full list of operation types in Appendix Table~\ref{tab:app-sft-data}. Applying $u$ produces an updated skill bank $\mathcal{B}'=\mathrm{Update}(\mathcal{B},u)$.

The objective is to optimize $M_\theta$ so that the updated skill bank improves the downstream performance of the frozen policy $\pi$ when used as prior knowledge. We write this objective as
\[
\begin{aligned}
\max_{\theta}\quad &
\mathbb{E}_{\tau,x'}\!\left[
R_{\mathrm{task}}\!\left(\pi(x' \mid \mathcal{B}')\right)
\right], \\
&
u = M_\theta(\tau,\mathcal{C}), \quad
\mathcal{B}' = \mathrm{Update}(\mathcal{B},u).
\end{aligned}
\]

where $R_{\mathrm{task}}$ denotes downstream task performance on future task $x'$.

\subsection{Skill Management Loop}

Figure~\ref{fig:pipeline} provides an overview of the skill-management framework of \Name. In this section, we describe how \Name organizes the skill bank and manages it through three major components, including skill extraction, skill evolution, and skill-bank maintenance, together with their corresponding operations.

% \paragraph{Multi-granularity Skill Bank.}
% \Name maintains a skill bank $\mathcal{B}$ composed of reusable procedural skills. Each skill is a markdown file including a title, a trigger condition describing when it should be applied, and instructions specifying the workflow, constraints, and actionable rules for the agent to follow. \yiran{Maybe a case and some rationale?} We organize the bank with two granularities. Task-level skills capture high-level strategies for solving a task or a family of tasks with similar goals. They are often distilled from trajectories that solve related queries and provide task-oriented guidance, such as how to inspect the repository, localize the issue, or validate a fix. In contrast, event-driven skills provide local guidance for recurring events during agent execution, such as command failures, error messages, test-output patterns, or repeated failure modes after specific actions. An event-driven skill specifies how the agent should react when the corresponding event is triggered, and can transfer across tasks because similar execution events recur in different software-engineering problems. This design supports both task-oriented solution strategies and agent-oriented execution rules within a unified skill bank.

\paragraph{Multi-granularity Skill Bank.}
\Name maintains a skill bank $\mathcal{B}$ composed of reusable procedural skills. Following prior work~\citep{ni2026trace2skill,xia2026skillrl,yang2026autoskill,han2026sweskillsbench}, each skill is represented as a markdown instruction file with a title, a trigger condition, and actionable instructions for the agent. Detailed skill representations and examples are provided in Appendix~\ref{app:skill-examples}. To cover both task-level procedural knowledge and reusable guidance for local execution events, we organize the bank with two granularities. Task-level skills capture high-level strategies for a task or a family of related tasks, such as how to inspect the repository, localize the issue, or validate a fix. They are often distilled from trajectories that solve related queries. Event-driven skills instead provide local guidance for recurring execution events, such as command failures, error messages, test-output patterns, or repeated failure modes after specific actions. They specify how the agent should react when the corresponding event is triggered, and can transfer across tasks because similar execution events recur across software-engineering problems.

\paragraph{Skill Bank Construction.}
For skill extraction, the input is trajectory evidence ranging from a single trajectory to a small set of related trajectories, and the output is either a new task-level skill, a new event-driven skill, or \texttt{skip} when the evidence does not support reusable knowledge. Task-level extraction abstracts high-level procedures from related task trajectories, while event-driven extraction captures local execution patterns from a specific trajectory. For skill evolution, the input is an existing skill together with new or failed trajectory evidence, and the output is a revised candidate skill or \texttt{skip}. This operation updates the applicability condition or procedural guidance of a skill when new evidence reveals missing cases, more effective procedures, or failure modes.

\paragraph{Skill Bank Maintenance.}
To keep the skill bank compact while allowing it to accumulate useful procedural knowledge over time, each newly extracted or evolved candidate skill is further passed to a maintenance stage. Similar skills are retrieved from the current bank and provided to $M_\theta$ together with the candidate. Based on this context, $M_\theta$ outputs a maintenance operation that either adds the candidate, merges it with an existing skill, or drops it. The \texttt{add} operation inserts the candidate as a new skill when it provides useful knowledge not covered by existing skills. The \texttt{merge} operation combines the candidate with an existing skill when they overlap but contain complementary guidance. The \texttt{drop} operation rejects candidates that are redundant, weakly grounded, overly specific, or unlikely to transfer.

\paragraph{Skill Usage.}
For downstream task solving, relevant skills are retrieved from the updated skill bank and inserted into the prompt of the frozen coding policy as prior knowledge. Retrieval is based on dense semantic similarity, matching task-level skills to the task goal and event-driven skills to the agent's current reasoning and observation through their trigger conditions. The original task prompt and policy parameters remain unchanged.

\subsection{Model Training}
\label{sec:3.3}

\begin{figure*}[t]
    \centering
    \includegraphics[width=0.98\textwidth]{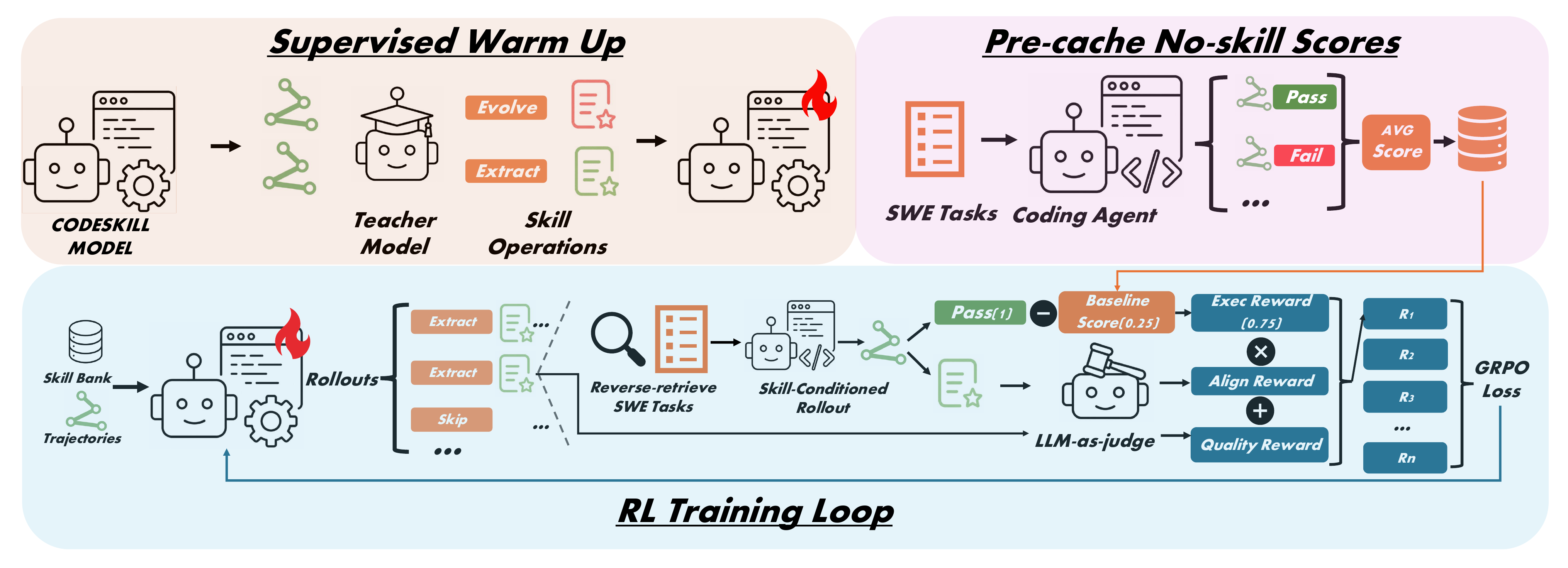}
    \vspace{-8pt}
    \caption{Training pipeline of \Name.}
    \label{fig:training}
    \vspace{-8pt}
\end{figure*}

Figure~\ref{fig:training} summarizes the training procedure of \Name. We first warm up $M_\theta$ on teacher-generated supervised data for skill extraction, evolution, and maintenance, and then optimize it with reinforcement learning using feedback from both rubric-based LLM-as-judge rewards and a frozen downstream coding policy.

\subsubsection{Supervised Warmup}
We first warm up $M_\theta$ with supervised data generated by teacher models. Given normalized coding-agent trajectories, optional existing skills, and retrieved skill-bank context, the teacher produces structured targets $u^\ast=(a^\ast,z^\ast)$ for skill extraction, evolution, and maintenance. Each example is written as $(q,u^\ast)$, where $q=(\tau,\tilde{s},\mathcal{C})$ contains trajectory evidence $\tau$, an optional candidate or existing skill $\tilde{s}$, and retrieved similar skills $\mathcal{C}\subseteq\mathcal{B}$. We fine-tune the base model with
\[
\mathcal{L}_{\mathrm{SFT}}
= - \mathbb{E}_{(q,u^\ast)}
\log M_\theta(u^\ast \mid q).
\]
This produces an initial policy that follows the operation schema before reinforcement learning.

\subsubsection{RL for \Name}

\paragraph{Attribution-Aware Hybrid Reward Design.}
To train $M_\theta$ with RL, we first introduce a rubric-based LLM-as-judge quality reward $R_Q \in [0,1]$ to score each generated operation and its content. For skills, the rubric evaluates grounding, reusability, specificity, format, and actionability, with action-specific criteria detailed in Appendix \ref{appen:reward}. However, quality reward alone is not execution-grounded, since a skill may look plausible but remain unused or ineffective.

We therefore define an execution reward by comparing the frozen coding policy's performance with and without the generated skill. Given a sampled operation $u$ that generates a valid skill $s_u$, we select a task instance from the RL training pool by reverse retrieval, $x_u \sim \mathrm{TopK}(s_u, \mathcal{D}_{\mathrm{task}})$, where the skill content retrieves task goals or baseline trajectories and $x_u$ is sampled from the top $K$ results. The no-skill baseline is
\[
b_\pi(x_u)=\frac{1}{n}\sum_{i=1}^{n} V(\tau_{\pi,i}^{0}),
\quad \tau_{\pi,i}^{0} \sim \pi(\cdot \mid x_u).
\]
We then run a skill-conditioned rollout $\tau^u_\pi \sim \pi(\cdot \mid x_u,s_u)$ and define the execution reward as the verifier improvement over the no-skill baseline, $R_E(u;x_u,\pi)=V(\tau^u_\pi)-b_\pi(x_u)$. Here $\tau_{\pi,i}^{0}$ is the $i$-th no-skill rollout, $\tau^u_\pi$ is the rollout conditioned on the skill produced by $u$, and $V(\tau)\in[0,1]$ is the task verifier score.
Execution improvement alone does not establish whether the generated skill contributed to the outcome: the coding policy may succeed without following the skill. We therefore use an alignment factor $R_A(u;\tau^u_\pi)\in[0,1]$ to judge whether the rollout actually matches the skill trigger condition and follows its workflow or constraints, with LLM-as-judge rubrics detailed in Appendix \ref{appen:reward}. The final reward for skill-output operations is
\[
R(u;q)=\lambda R_Q(u;q)+R_A(u;\tau^u_\pi) R_E(u;x_u,\pi)
\]
This attribution-aware design combines dense quality supervision with baseline-relative execution feedback, gated by behavioral evidence of skill use. It provides a practical proxy for skill-level credit assignment by discounting execution feedback when the rollout does not reflect the skill's trigger conditions or prescribed behavior.

% \paragraph{Policy Optimization Loop.}
% Algorithm \ref{alg:rl-training} summarizes the RL loop. For each prompt $q$, we sample a group of $G$ operations $\{u_j\}_{j=1}^{G}$ from the old policy $M_{\theta_{\mathrm{old}}}$. Each operation is parsed and scored with the LLM-as-judge quality reward. For operations that produce an injectable skill, including extracted, evolved, and merged skills, we retrieve an evaluation instance, run the frozen coding policy $\pi$ with the skill, and compute the execution and alignment rewards. Let $R_j=R(u_j;q)$ denote the final reward of $u_j$. We update $M_\theta$ with the GRPO objective
% \[
% \mathcal{L}_{\mathrm{GRPO}}
% = -\frac{1}{G}\sum_{j=1}^{G}
% \left[\min(\rho_j \hat{A}_j,\bar{\rho}_j \hat{A}_j)-\beta D_j\right].
% \]
% Here $\rho_j=M_\theta(u_j\mid q)/M_{\theta_{\mathrm{old}}}(u_j\mid q)$, $\bar{\rho}_j=\mathrm{clip}(\rho_j,1-\epsilon,1+\epsilon)$, $\hat{A}_j=(R_j-\mu_R)/(\sigma_R+\epsilon_A)$ is the group-normalized advantage, $\mu_R$ and $\sigma_R$ are the mean and standard deviation of rewards within the group, and $D_j$ is the KL penalty to the reference model~\citep{deepseekmath2024grpo}.

\paragraph{Policy Optimization Loop.}
Algorithm \ref{alg:rl-training} summarizes the RL loop. For each prompt $q$, we sample a group of $G$ operations $\{u_j\}_{j=1}^{G}$ from the old policy $M_{\theta_{\mathrm{old}}}$. Each operation $u_j$ is parsed and assigned a quality reward $R_Q(u_j;q)$ by the LLM-as-judge. For operations that produce an injectable skill $s_j$, including extracted, evolved, and merged skills, we reverse-retrieve a task instance $x_j$ from the RL training pool and run the frozen coding policy to obtain a skill-conditioned trajectory $\tau_j \sim \pi(\cdot \mid x_j,s_j)$. We compute the execution reward $R_E=V(\tau_j)-b_\pi(x_j)$ relative to the pre-computed no-skill baseline, and the alignment factor $R_A(u_j;\tau_j)$ from evidence of skill use in the trajectory. The final reward for each skill-output operation is then $R_j=\lambda R_Q(u_j;q)+R_A(u_j;\tau_j)R_E$. For \texttt{skip}, \texttt{add}, and \texttt{drop}, we use only the quality reward with a higher weight. After assigning rewards to all operations in the group, we normalize them into advantages and update $M_\theta$ with the GRPO objective
\[
\mathcal{L}_{\mathrm{GRPO}}
= -\frac{1}{G}\sum_{j=1}^{G}
\left[
\min(\rho_j \hat{A}_j,\bar{\rho}_j \hat{A}_j)
-\beta D_j
\right].
\]
Here $\rho_j=M_\theta(u_j\mid q)/M_{\theta_{\mathrm{old}}}(u_j\mid q)$, $\bar{\rho}_j=\mathrm{clip}(\rho_j,1-\epsilon,1+\epsilon)$, $\hat{A}_j=(R_j-\mu_R)/(\sigma_R+\epsilon_A)$ is the group-normalized advantage, $\mu_R$ and $\sigma_R$ are the mean and standard deviation of rewards within the group, and $D_j$ is the KL penalty to the reference model~\citep{deepseekmath2024grpo}.

\begin{algorithm}[t]
\caption{GRPO Training for \Name}
\label{alg:rl-training}
\footnotesize
\algrenewcommand\algorithmicindent{1.0em}
\begin{algorithmic}[1]
\Require prompts $\mathcal{Q}$, task pool $D$, downstream policy $\pi$,
verifier $V$, old policy $M_o$

\State $\tau_i^0(x)\sim\pi(\cdot\mid x),\quad
x\in D,\ i=1,\ldots,n$
\Comment{baseline rollouts}

\State $b_\pi(x)\gets n^{-1}\sum_{i=1}^{n}V(\tau_i^0(x)),
\quad x\in D$
\Comment{pre-cache scores}

\For{$q\in\mathcal{Q}$}
    \State $\{u_j\}_{j=1}^{G}\sim M_o(\cdot\mid q)$
    \Comment{\Name rollout}

    \For{$j=1,\ldots,G$}
        \State $R_Q\gets R_Q(u_j;q);\quad
        s_j\gets\mathrm{Skill}(u_j)$
        \Comment{compute quality score; extract skill}

        \State $x_j\sim\mathrm{TopK}(s_j,D);\quad
        \tau_j\sim\pi(\cdot\mid x_j,s_j)$
        \Comment{reverse retrieval; downstream policy execution}

        \State $R_E\gets V(\tau_j)-b_\pi(x_j);\quad
        R_A\gets R_A(u_j;\tau_j)$
        \Comment{execution reward; alignment reward}

        \State $R_j\gets\lambda R_Q+R_A R_E$
        \Comment{final hybrid reward}
    \EndFor

    \State $\hat{A}_j\gets
    (R_j-\mu_R)/(\sigma_R+\epsilon_A),\ j=1,\ldots,G$;
    update $M_\theta$ with $\mathcal{L}_{\mathrm{GRPO}}$
    \Comment{advantage; policy update}
\EndFor
\end{algorithmic}
\end{algorithm}

\textbf{Three-stage Curriculum Training.} We train \Name with a three-stage curriculum that follows the life cycle of a skill bank.
Skill management forms a natural data loop, extracted skills become candidates for later evolution, both extracted and evolved skills become candidates for maintenance, and maintenance decisions determine what remains in the bank for future updates. 
Since \Name must learn multiple operation types under sparse verifiable feedback, we introduce them progressively so that earlier-stage skill outputs can be reused as training context for later-stage evolution and maintenance. \ding{182} trains skill extraction, teaching $M_\theta$ to generate task-level and event-driven skills from trajectory evidence. \ding{183} adds skill evolution, where existing skills are revised using new or failed trajectory evidence. \ding{184} trains the full pipeline by adding skill-bank maintenance, so that extracted and evolved candidates can be added, merged, or dropped according to the current bank context. This curriculum aligns the full skill-management loop, and the implementation details are provided in Appendix~\ref{app:training}.
\section{Experiments}
\subsection{Experimental Setup}
\paragraph{Implementation Details.}
We instantiate \Name by initializing $M_\theta$ from Qwen3.5-4B~\citep{qwen3technicalreport2025}. For both supervised warmup and RL, we collect coding-agent trajectories from SWE-Bench Verified and SWE-smith using mini-SWE-agent, and from EnvBench using a ReAct-style bash agent ~\citep{swebench2024,swebenchverified2024,swesmith2025,minisweagent2025,sweagent2024,envbench2025,react2023}. GPT-5.4-mini is used as the teacher model to generate supervised targets from trajectory evidence and skill-bank context for skill extraction and skill-bank maintenance~\citep{openai2026gpt54mini}. Skill evolution data is synthesized by pairing existing skills with related trajectory evidence, avoiding an additional skill-conditioned rollout for revised-skill supervision. During RL, the frozen downstream coding policy $\pi$ and no-skill baseline rollouts use Qwen3.5-35B-A3B. GPT-5.4-mini is also used as the LLM-as-judge for quality and alignment rewards. More implementation details are provided in Appendix \ref{app:training}.

\paragraph{Benchmarks.}
We evaluate \Name on three software-engineering benchmarks. EnvBench tests environment setup and dependency repair, SWE-Bench Verified tests repository-level issue resolution, and Terminal-Bench 2 tests terminal-based problem solving~\citep{envbench2025,swebench2024,swebenchverified2024,terminalbench2026}. Since EnvBench and SWE-Bench Verified are also used for training, we use disjoint held-out splits for evaluation. For SWE-Bench Verified, we randomly sample 150 of the 500 Verified instances for evaluation and use the remaining 350 for RL training. For EnvBench, we sample 150 of 994 repositories for evaluation, covering 107 JVM and 43 Python repositories, and split the remaining 844 repositories evenly for SFT trajectory collection and RL training. Terminal-Bench 2 is not used during training, and serves as an out-of-distribution benchmark for testing whether the learned skill-management policy generalizes to new coding-agent tasks. We use mini-SWE-agent for SWE-Bench Verified and Terminal-Bench 2, and a ReAct-style agent for EnvBench~\citep{sweagent2024,react2023}. Although the benchmarks and agent wrappers differ, all agents interact with the environment through bash-command-based actions. We therefore normalize their trajectories into the same reasoning-action-observation format, allowing a single \Name model to manage skills across different SWE tasks. We report task pass rate on all benchmarks. For EnvBench, we also report average issue count, and for all benchmarks we report the average number of reasoning steps on the common subset
of instances solved by all compared methods within each benchmark
and downstream-policy setting.

\paragraph{Baselines.}
We compare \Name with representative memory and skill-management baselines. First, we evaluate prompt-based skill-management pipelines that use the same operation space as \Name but replace the learned policy with fixed prompts and heuristic decisions. This baseline follows the common design of prior automatic skill-bank management methods, such as SkillRL and AutoSkill~\citep{yang2026autoskill,xia2026skillrl}, while keeping the downstream coding policy frozen. We also compare with reasoning-oriented subtask-level memory, a strong coding-agent memory baseline that decomposes long software-engineering trajectories into subtasks and distills reusable memories from the resulting sub-trajectories~\citep{shen2026subtaskmemory}.

\paragraph{Test-Time Skill Bank Construction.}
We construct the skill bank through two preparatory passes over the test set before final evaluation. In the first pass, \Name extracts skills from no-skill rollout trajectories. In the second pass, it evolves skills using trajectories from skill-conditioned rollouts. Every extracted or evolved candidate undergoes maintenance through add, merge, or drop decisions. Together, these passes simulate a skill lifecycle of experience accumulation and refinement across tasks. We track the source instances of each skill, preserving their provenance through evolution and merging. During final evaluation, we exclude any skill whose provenance includes the current test instance to prevent access to knowledge derived from that instance. Task-level and event-driven skills are retrieved using task goals and local execution signals, respectively. Detailed retrieval procedures are provided in Appendix \ref{app:exp-details}.

\begin{table*}[t]
\centering
\scriptsize
\setlength{\tabcolsep}{2.0pt}
\renewcommand{\arraystretch}{1.08}
\caption{Main results on SWE-related benchmarks. Success is task pass rate. Issues denotes the average issue count on EnvBench, and Steps denotes the average reasoning steps on solved instances. Avg. reports macro-average success and reasoning steps across benchmarks.}
\label{tab:main-results}
\resizebox{0.95\textwidth}{!}{
\begin{tabular}{llcccccccccc|cc}
\toprule
\multirow{2}{*}{Method} & \multirow{2}{*}{Skill Backbone}
& \multicolumn{3}{c}{EnvBench-Python}
& \multicolumn{3}{c}{EnvBench-Java}
& \multicolumn{2}{c}{SWE-Bench-Verified}
& \multicolumn{2}{c|}{Terminal-Bench 2}
& \multicolumn{2}{c}{Average} \\
\cmidrule(lr){3-5}
\cmidrule(lr){6-8}
\cmidrule(lr){9-10}
\cmidrule(lr){11-12}
\cmidrule(lr){13-14}
& & Succ. $\uparrow$ & Issues $\downarrow$ & Steps $\downarrow$
& Succ. $\uparrow$ & Issues $\downarrow$ & Steps $\downarrow$
& Succ. $\uparrow$ & Steps $\downarrow$
& Succ. $\uparrow$ & Steps $\downarrow$
& Succ. $\uparrow$ & Steps $\downarrow$ \\
\midrule
\multicolumn{12}{c|}{\textit{Frozen coding policy: Qwen3.5-35B-A3B}}
& \multicolumn{2}{c}{} \\
\midrule
No-skill baseline & -- & 6.98 & 81.70 & 30.00 & 27.10 & 6.10 & 14.03  & 57.33 & 88.41 & 25.88 & 44.05 & 29.32 & 44.12 \\
Subtask Memory & GPT-5.4-mini &  9.30 & 86.33 & 26.50 & 32.71 & \textbf{4.25} & 13.34 & 61.33 & 81.31 & 30.59 & 37.88 & 33.48 & 39.76 \\
Prompt Skill Mgmt. & Qwen3.5-4B & 4.65 & 91.95 & 26.80 & 30.84 & 4.67 & \textbf{13.09} & 58.67 & 76.69 & 24.71 & 39.73& 29.72 & 39.08 \\
Prompt Skill Mgmt. & GPT-5.4-mini & 11.63 & \textbf{31.58} & 25.00 & 36.45 &  5.49& 13.72 & 64.67 & 75.97 & 28.24 & 33.38& 35.25 & 37.02\\
\cdashline{1-14}
SFT-\Name & Qwen3.5-4B & 13.95& 88.60 & 25.25 & 35.51 &  7.85&13.65 & 64.00 & 78.65 & 24.71 & \textbf{34.62} & 34.54 & 38.04 \\
\Name & Qwen3.5-4B & \textbf{18.60} & 62.74 & \textbf{24.33} & \textbf{38.32} & 5.35 &  14.00& \textbf{68.00} & \textbf{70.94} & \textbf{36.47} & 34.69 & \textbf{40.35} & \textbf{35.99} \\
\midrule
\multicolumn{12}{c|}{\textit{Frozen coding policy: GPT-5.4-mini}}
& \multicolumn{2}{c}{} \\
\midrule
No-skill baseline & - &4.65 & 70.72 & 13.00 & 15.89 & 3.62 & 9.76 & 46.67 & 7.56 & 20.00 & 8.23 & 21.80 & 9.64 \\
Subtask Memory & GPT-5.4-mini & 9.30 & 66.77 & 11.81 & 25.23 & 3.49 & 9.64 & 52.67 & 6.73 & 22.35 & \textbf{6.06} & 27.39 & 8.56 \\
Prompt Skill Mgmt. & Qwen3.5-4B & 6.98 & 31.19 & 10.67 & 22.43 & 2.56 & 9.25 & 50.67 & \textbf{6.32} & 23.53 & 6.25 & 25.90 & 8.12 \\
Prompt Skill Mgmt. & GPT-5.4-mini & 9.30 & 79.60 & 10.50 & 24.30 & 2.27 & 9.35 & \textbf{56.67} & 6.53 & 21.18 & \textbf{6.06} & 27.86 & 8.11 \\
\cdashline{1-14}
SFT-\Name & Qwen3.5-4B & 6.98 & 70.77 & 10.50 & 23.36 & 2.21 & \textbf{8.25} & 54.67 & 6.66 & 23.53 & 6.50 & 27.14 & \textbf{7.98} \\
\Name & Qwen3.5-4B & \textbf{13.95} & \textbf{25.05}& \textbf{9.00}& \textbf{27.10} & \textbf{1.79} & 9.10 & 56.00 & 6.43 & \textbf{29.41} & 7.58 & \textbf{31.62} & 8.03 \\
\bottomrule
\end{tabular}
}
\end{table*}

\subsection{Main Results}
Table~\ref{tab:main-results} reports the main evaluation results across EnvBench, SWE-Bench Verified, and Terminal-Bench 2. We compare \Name with no-skill execution, subtask-level memory, and prompt-based skill management using different skill backbones, including open-source Qwen3.5-4B and closed-source GPT-5.4-mini. The table also reports results under two frozen downstream coding policies, Qwen3.5-35B-A3B and GPT-5.4-mini.

% \paragraph{Comparison with Baselines.}
% We first compare pass rates under Qwen3.5-35B-A3B as the downstream coding policy. Compared with the no-skill baseline, \Name improves average pass rate from 29.57 to 39.26, showing that managed procedural skills provide substantial benefit to downstream coding agents. In contrast, prompt-based skill management with the small Qwen3.5-4B backbone brings little improvement over no-skill, indicating that simply applying a weak model to the same skill-management pipeline is insufficient. Stronger memory and skill baselines are more effective, as subtask-level memory and prompt-based skill management with GPT-5.4-mini improve average pass rate over no-skill by 3.91 and 5.68, respectively. However, they still underperform \Name by 5.78 and 4.01 average pass rate. SFT-\Name also improves consistently over no-skill, showing that supervised warmup learns useful skill-management behavior, but it remains weaker than prompt-based management with a stronger closed-source model. After RL training, \Name achieves the best pass rate on all benchmark groups, improving over the strongest baseline by 11.4\% relative gain on average. This suggests that downstream-feedback optimization learns a skill-management policy better suited to the frozen coding agent than fixed-prompt skill distillation methods. 

\paragraph{Comparison with Baselines.}
We first compare \Name with baselines under Qwen3.5-35B-A3B as the downstream coding policy. Compared with the no-skill baseline, \Name improves average pass rate from 29.32 to 40.35, showing the benefit of managed procedural skills for downstream coding agents. Prompt-based skill management with the small Qwen3.5-4B backbone brings little improvement over no-skill, indicating that a weak model is insufficient for the same skill-management pipeline. Stronger baselines are more effective, with subtask-level memory and GPT-5.4-mini prompt-based skill management improving over no-skill by 4.16 and 5.93 average pass rate, respectively. However, they still underperform \Name by 6.87 and 5.10. SFT-\Name improves average pass rate over no-skill, but remains weaker than prompt-based management with a stronger closed-source model. After RL training, \Name achieves the best pass rate on all benchmark groups, improving over the strongest baseline by 14.5\% relative gain on average. Overall, the results suggest that downstream-feedback optimization learns a skill-management policy better suited to the frozen coding agent than fixed-prompt skill distillation methods.

\begin{table*}[t]
\centering
\scriptsize
\setlength{\tabcolsep}{2.5pt}
\renewcommand{\arraystretch}{1.08}
\caption{Ablation study of the skill-bank lifecycle with Qwen3.5-35B-A3B as the frozen coding policy. Skill Num. denotes the number of skills in the resulting skill bank.}
\label{tab:lifecycle-ablation}
\begin{tabular}{lccccccccccc}
\toprule
\multirow{2}{*}{Variant}
& \multicolumn{3}{c}{EnvBench-Python}
& \multicolumn{3}{c}{EnvBench-Java}
& \multicolumn{2}{c}{SWE-Bench Verified}
& \multicolumn{2}{c}{Terminal-Bench 2}
& \multirow{2}{*}{Skill Num.} \\
\cmidrule(lr){2-4}
\cmidrule(lr){5-7}
\cmidrule(lr){8-9}
\cmidrule(lr){10-11}
& Succ. $\uparrow$ & Issues $\downarrow$ & Steps $\downarrow$
& Succ. $\uparrow$ & Issues $\downarrow$ & Steps $\downarrow$
& Succ. $\uparrow$ & Steps $\downarrow$
& Succ. $\uparrow$ & Steps $\downarrow$
& \\
\midrule
No-skill baseline & 6.98 & 81.70 & 30.00 & 27.10 & 6.10 & 14.03 & 57.33 & 88.41 & 25.88 & 44.05 & 0 \\
Event-driven only & 16.28 & 53.98 & 26.57 & 32.71 & 5.65 & \textbf{13.57} & 62.00 & 76.99 & 29.41 & 33.83 & 916 \\
Task-level only & 11.63 & 78.33 & 26.60 & 30.84 & \textbf{3.28} & 14.57 & 58.67 & 88.25 & 28.24 & 46.25 & 336 \\
Extraction only & 13.95 & 58.12 & 25.09 & \textbf{41.12} & 5.79 & 14.15 & 64.67 & 78.05 & 31.76 & \textbf{30.63} & 1252 \\
Extraction + Evolution & \textbf{18.60} & \textbf{43.00} & 24.74 & 39.25 & 4.16 & 13.90 & 67.33 & 77.28 & 32.94 & 32.17 & 1252 \\
Full lifecycle & \textbf{18.60} & 62.74 & \textbf{24.33} & 38.32 & 5.35 & 14.00 & \textbf{68.00} & \textbf{70.94} & \textbf{36.47} & 34.69 & 676 \\
\bottomrule
\end{tabular}
\end{table*}

\paragraph{Efficiency.}
We also compare the average number of reasoning steps on the common subset of instances successfully solved by all compared methods, ensuring that all averages are computed over the same tasks. Skill-based methods generally reduce the reasoning cost compared with the no-skill baseline, suggesting that retrieved procedural guidance can help the coding agent reach successful solutions more directly. Under the Qwen3.5-35B-A3B policy, \Name achieves the lowest average reasoning steps, reducing the average from 44.12 for no-skill to 35.99. This indicates that learned skill management improves not only task success but also the efficiency of successful problem solving.

\paragraph{Generalization across Tasks.}
We first examine whether \Name generalizes beyond the task distributions used for training. \Name is trained with trajectories from EnvBench and SWE-Bench Verified, while Terminal-Bench 2 is held out and used only for evaluation. Although Terminal-Bench 2 shares the same bash-command interaction interface, it contains different terminal-based problem-solving tasks and therefore tests out-of-distribution generalization within software engineering. Under the Qwen3.5-35B-A3B coding policy, \Name improves Terminal-Bench 2 pass rate from 25.88 to 36.47 over no-skill, and also outperforms the strongest baseline on this benchmark, subtask-level memory with GPT-5.4-mini, by 5.88 pass rate. In contrast, SFT-\Name does not improve over no-skill on Terminal-Bench 2. This suggests that supervised warmup alone may not generalize reliably to unseen SWE task types, while RL with downstream feedback helps \Name learn more transferable skill-management behavior.

\paragraph{Generalization across Policies.}
We further evaluate whether the learned skill-management policy transfers to a different downstream coding agent. Although \Name is optimized with reward rollouts from Qwen3.5-35B-A3B, we also use GPT-5.4-mini as the frozen coding policy while keeping the same skill-management and retrieval protocol. As shown in Table~\ref{tab:main-results}, \Name again achieves the best average pass rate, improving over the no-skill baseline by 9.82 and over the strongest prompt-based or memory baseline by 3.76, corresponding to relative gains of 45.0\% and 13.5\%, respectively. This suggests that \Name does not merely overfit to the downstream policy used during training, but learns skill-management behavior that can benefit different coding models.

\subsection{Ablation \& Analysis}
\paragraph{Lifecycle Ablation.}
To understand how each component of the skill-bank lifecycle affects downstream coding agents, we conduct the ablation study in Table~\ref{tab:lifecycle-ablation}. We fix Qwen3.5-35B-A3B as the downstream coding policy and compare variants that progressively enable different skill granularities and lifecycle operations. The event-driven-only and task-level-only variants use a single skill granularity, while extraction only combines both. Event-driven and task-level skills each improve over the no-skill baseline on different tasks, and their combination further improves average performance, suggesting that local execution guidance and high-level task strategies capture complementary knowledge. Adding skill evolution improves the average pass rate from 37.88 to 39.53 over extraction only. Full lifecycle maintenance further improves the average pass rate to 40.35 while shrinking the skill bank from 1252 to 676 skills, a reduction of approximately 46\%. These results suggest that maintenance improves the compactness of the skill bank without sacrificing aggregate performance. The additional performance gain may partly arise from reducing redundant or conflicting skills, making retrieved guidance less repetitive and more consistent. Finally, all skill-based variants reduce average reasoning steps over the no-skill baseline, suggesting that procedural skill guidance contributes to more efficient successful task execution.

\begin{figure*}[t]
\vspace{-8pt}
    \centering
    \includegraphics[width=0.98\textwidth]{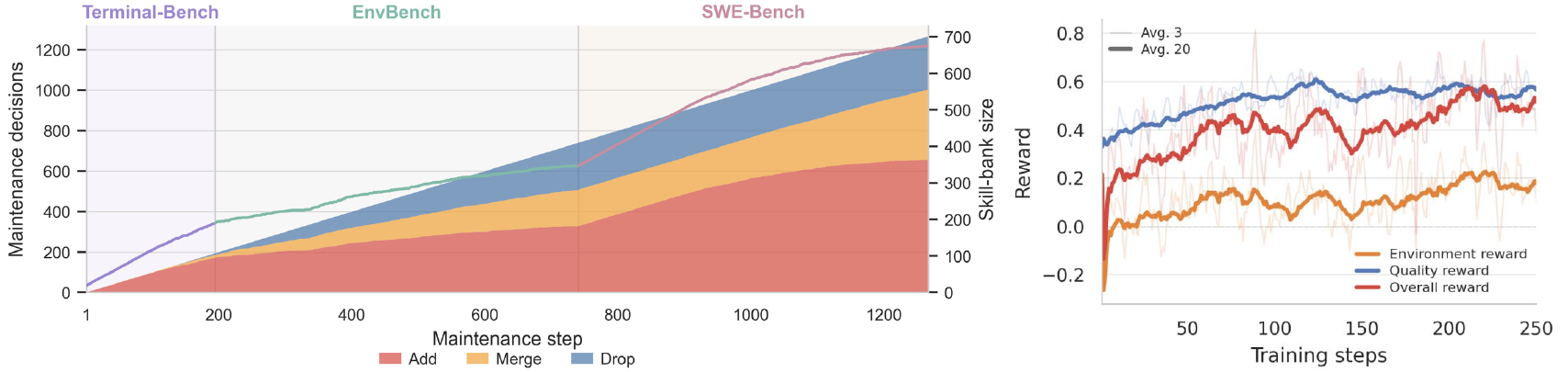}
    \caption{
    Analysis of skill-bank maintenance and RL training dynamics.
    The left panel shows cumulative add, merge, and drop decisions together with the resulting skill-bank size during skill-bank maintenance.
    The right panel shows the reward number over the stage-3 training process.
    }
    \label{fig:ablation-analysis}
    \vspace{-8pt}
\end{figure*}

\paragraph{Skill-Bank Maintenance Dynamics.}
Figure~\ref{fig:ablation-analysis} visualizes skill-bank maintenance during test-time skill construction. As \Name processes evaluation instances, each newly extracted or evolved candidate skill enters the maintenance stage, where it can be added, merged with an existing skill, or dropped. The figure records cumulative maintenance decisions and the resulting skill-bank size. Add decisions dominate early, when the bank is small and most candidates provide uncovered knowledge. As the bank grows, merge and drop decisions become more frequent, indicating that \Name identifies overlapping, redundant, or low-value candidates rather than adding every candidate to the bank.
The benchmark-wise dynamics illustrate how maintenance behavior varies with the amount of task experience within each benchmark. Terminal-Bench has fewer test instances and candidate skills, and add decisions remain prominent in its segment. EnvBench and SWE-Bench provide more opportunities to accumulate related skills, with merge and drop decisions consolidating overlapping knowledge. We hypothesize that, under a relatively stable task distribution, continued evolution could drive the bank toward a steady size as reusable patterns become increasingly covered. Meanwhile, add decisions remain active across benchmark segments, consistent with the need to incorporate new procedural knowledge when the task distribution changes.

\paragraph{Reward Optimization Dynamics.}
The right panel of Figure~\ref{fig:ablation-analysis} shows reward dynamics during phase-3 RL training. Unlike directly optimizing a task-solving policy, \Name affects outcomes only after its skills are retrieved, injected, and followed in long-horizon software tasks, making the environment reward noisy. Still, its 20-step trend increases from 0.004 in steps 1--20 to 0.158 in steps 180--200, suggesting that \Name gradually learns management decisions that improve the downstream agent over its no-skill baseline. The quality reward, provided by rubric-based LLM-as-judge feedback, is more stable and rises before stabilizing after roughly 100 steps. The overall reward follows the same upward pattern, indicating that the hybrid objective improves both executable utility and skill quality during RL training.

\section{Conclusion}
\vspace{-5pt}
We presented \Name, an LLM-based framework that learns to extract, evolve, and maintain reusable skills from coding-agent trajectories. \Name formulates skill management as a learnable policy and optimizes it with verifiable downstream feedback and rubric-based skill-quality judgments by reinforcement learning. Experiments on EnvBench, SWE-Bench Verified, and Terminal-Bench 2 show that learned skill management improves over no-skill execution, prompt-based skill management, and memory-based baselines. These results highlight learned procedural-memory management as a promising direction for coding agents that accumulate and reuse long-horizon software-engineering experience. These gains are achieved without updating the downstream coding policy, emphasizing the value of learning how experience is organized and reused. Future work could extend this framework to strictly streaming task settings and evaluate the end-to-end cost of skill acquisition, maintenance, and use.
\subsection*{AI use statement}
We used generative AI tools to polish the language and improve the readability of the manuscript. As part of the research pipeline, we used LLMs to generate supervised skill-management targets and to provide rubric-based quality and behavior-alignment judgments, as detailed in Section \ref{sec:3.3} and Appendices \ref{app:training}--\ref{appen:reward}. The authors reviewed and edited the AI-assisted text and take full responsibility for the final content, claims, and reported results.
\bibliography{iclr2027_conference}

@misc{chen2025sweexp,
  title = {{SWE-Exp}: Experience-Driven Software Issue Resolution},
  author = {Silin Chen and Shaoxin Lin and Yuling Shi and Heng Lian and Xiaodong Gu and Longfei Yun and Dong Chen and Lin Cao and Jiyang Liu and Nu Xia and Qianxiang Wang},
  year = {2025},
  eprint = {2507.23361},
  archivePrefix = {arXiv},
  primaryClass = {cs.SE},
  doi = {10.48550/arXiv.2507.23361},
  url = {https://arxiv.org/abs/2507.23361}
}

@inproceedings{tang2025agentkb,
  title = {{AGENT KB}: A Hierarchical Memory Framework for Cross-Domain Agentic Problem Solving},
  author = {Xiangru Tang and Tianrui Qin and Tianhao Peng and Ziyang Zhou and Yanjun Shao and Tingting Du and Xinming Wei and He Zhu and Ge Zhang and Jiaheng Liu and Xingyao Wang and Sirui Hong and Chenglin Wu and Wangchunshu Zhou},
  booktitle = {ICML 2025 Workshop on Collaborative and Federated Agentic Workflows},
  year = {2025},
  note = {Oral},
  url = {https://openreview.net/forum?id=ohXoWHlrn8}
}

@inproceedings{ouyang2026reasoningbank,
  title = {{ReasoningBank}: Scaling Agent Self-Evolving with Reasoning Memory},
  author = {Siru Ouyang and Jun Yan and I-Hung Hsu and Yanfei Chen and Ke Jiang and Zifeng Wang and Rujun Han and Long T. Le and Samira Daruki and Xiangru Tang and Vishy Tirumalashetty and George Lee and Mahsan Rofouei and Hangfei Lin and Jiawei Han and Chen-Yu Lee and Tomas Pfister},
  booktitle = {International Conference on Learning Representations},
  year = {2026},
  url = {https://openreview.net/forum?id=jL7fwchScm}
}

@misc{zhu2026swecontextbench,
  title = {{SWE Context Bench}: A Benchmark for Context Learning in Coding},
  author = {Jiayuan Zhu and Junde Wu and Minhao Hu and Shengda Zhu and Jiazhen Pan and Weixiang Shen and Yijun Yang and Fenglin Liu and Jianye Hao and Yueming Jin and Qirong Ho and Min Xu},
  year = {2026},
  eprint = {2602.08316},
  archivePrefix = {arXiv},
  primaryClass = {cs.SE},
  doi = {10.48550/arXiv.2602.08316},
  url = {https://arxiv.org/abs/2602.08316}
}

@misc{shen2026subtaskmemory,
  title = {Structurally Aligned Subtask-Level Memory for Software Engineering Agents},
  author = {Kangning Shen and Jingyuan Zhang and Chenxi Sun and Wencong Zeng and Yang Yue},
  year = {2026},
  eprint = {2602.21611},
  archivePrefix = {arXiv},
  primaryClass = {cs.SE},
  doi = {10.48550/arXiv.2602.21611},
  url = {https://arxiv.org/abs/2602.21611}
}

@misc{han2026sweskillsbench,
  title = {{SWE-Skills-Bench}: Do Agent Skills Actually Help in Real-World Software Engineering?},
  author = {Tingxu Han and Yi Zhang and Wei Song and Chunrong Fang and Zhenyu Chen and Youcheng Sun and Lijie Hu},
  year = {2026},
  eprint = {2603.15401},
  archivePrefix = {arXiv},
  primaryClass = {cs.SE},
  doi = {10.48550/arXiv.2603.15401},
  url = {https://arxiv.org/abs/2603.15401}
}

@misc{chhikara2025mem0,
  title = {{Mem0}: Building Production-Ready AI Agents with Scalable Long-Term Memory},
  author = {Prateek Chhikara and Dev Khant and Saket Aryan and Taranjeet Singh and Deshraj Yadav},
  year = {2025},
  eprint = {2504.19413},
  archivePrefix = {arXiv},
  primaryClass = {cs.CL},
  doi = {10.48550/arXiv.2504.19413},
  url = {https://arxiv.org/abs/2504.19413}
}

@misc{wu2025evolver,
  title = {{EvolveR}: Self-Evolving {LLM} Agents through an Experience-Driven Lifecycle},
  author = {Rong Wu and Xiaoman Wang and Jianbiao Mei and Pinlong Cai and Daocheng Fu and Cheng Yang and Licheng Wen and Xuemeng Yang and Yufan Shen and Yuxin Wang and Botian Shi},
  year = {2025},
  eprint = {2510.16079},
  archivePrefix = {arXiv},
  primaryClass = {cs.CL},
  doi = {10.48550/arXiv.2510.16079},
  url = {https://arxiv.org/abs/2510.16079}
}

@misc{zhang2026memrl,
  title = {{MemRL}: Self-Evolving Agents via Runtime Reinforcement Learning on Episodic Memory},
  author = {Shengtao Zhang and Jiaqian Wang and Ruiwen Zhou and Junwei Liao and Yuchen Feng and Zhuo Li and Yujie Zheng and Weinan Zhang and Ying Wen and Zhiyu Li and Feiyu Xiong and Yutao Qi and Bo Tang and Muning Wen},
  year = {2026},
  eprint = {2601.03192},
  archivePrefix = {arXiv},
  primaryClass = {cs.CL},
  doi = {10.48550/arXiv.2601.03192},
  url = {https://arxiv.org/abs/2601.03192}
}

@inproceedings{xia2026skillrl,
  title = {{SkillRL}: Evolving Agents via Recursive Skill-Augmented Reinforcement Learning},
  author = {Peng Xia and Jianwen Chen and Hanyang Wang and Jiaqi Liu and Kaide Zeng and Yu Wang and Siwei Han and Yiyang Zhou and Xujiang Zhao and Haifeng Chen and Zeyu Zheng and Cihang Xie and Huaxiu Yao},
  booktitle = {ICLR 2026 Workshop on Memory Agents},
  year = {2026},
  note = {Oral},
  url = {https://openreview.net/forum?id=By7Pj576U3}
}

@misc{yang2026autoskill,
  title = {{AutoSkill}: Experience-Driven Lifelong Learning via Skill Self-Evolution},
  author = {Yutao Yang and Junsong Li and Qianjun Pan and Bihao Zhan and Yuxuan Cai and Lin Du and Jie Zhou and Kai Chen and Qin Chen and Xin Li and Bo Zhang and Liang He},
  year = {2026},
  eprint = {2603.01145},
  archivePrefix = {arXiv},
  primaryClass = {cs.AI},
  doi = {10.48550/arXiv.2603.01145},
  url = {https://arxiv.org/abs/2603.01145}
}

@misc{ni2026trace2skill,
  title = {{Trace2Skill}: Distill Trajectory-Local Lessons into Transferable Agent Skills},
  author = {Jingwei Ni and Yihao Liu and Xinpeng Liu and Yutao Sun and Mengyu Zhou and Pengyu Cheng and Dexin Wang and Erchao Zhao and Xiaoxi Jiang and Guanjun Jiang},
  year = {2026},
  eprint = {2603.25158},
  archivePrefix = {arXiv},
  primaryClass = {cs.AI},
  doi = {10.48550/arXiv.2603.25158},
  url = {https://arxiv.org/abs/2603.25158}
}

@misc{zhang2026coevoskills,
  title = {{CoEvoSkills}: Self-Evolving Agent Skills via Co-Evolutionary Verification},
  author = {Hanrong Zhang and Shicheng Fan and Henry Peng Zou and Yankai Chen and Zhenting Wang and Jiayu Zhou and Chengze Li and Wei-Chieh Huang and Yifei Yao and Kening Zheng and Xue Liu and Xiaoxiao Li and Philip S. Yu},
  year = {2026},
  eprint = {2604.01687},
  archivePrefix = {arXiv},
  primaryClass = {cs.AI},
  doi = {10.48550/arXiv.2604.01687},
  url = {https://arxiv.org/abs/2604.01687}
}

@misc{ma2026skillclaw,
  title = {{SkillClaw}: Let Skills Evolve Collectively with Agentic Evolver},
  author = {Ziyu Ma and Shidong Yang and Yuxiang Ji and Xucong Wang and Yong Wang and Yiming Hu and Tongwen Huang and Xiangxiang Chu},
  year = {2026},
  eprint = {2604.08377},
  archivePrefix = {arXiv},
  primaryClass = {cs.AI},
  doi = {10.48550/arXiv.2604.08377},
  url = {https://arxiv.org/abs/2604.08377}
}

@misc{zhang2026skillflow,
  title = {{SkillFlow}: Benchmarking Lifelong Skill Discovery and Evolution for Autonomous Agents},
  author = {Ziao Zhang and Kou Shi and Shiting Huang and Avery Nie and Yu Zeng and Yiming Zhao and Zhen Fang and Qisheng Su and Haibo Qiu and Wei Yang and Qingnan Ren and Shun Zou and Wenxuan Huang and Lin Chen and Zehui Chen and Feng Zhao},
  year = {2026},
  eprint = {2604.17308},
  archivePrefix = {arXiv},
  primaryClass = {cs.AI},
  doi = {10.48550/arXiv.2604.17308},
  url = {https://arxiv.org/abs/2604.17308}
}

@inproceedings{shinn2023reflexion,
  title = {Reflexion: Language Agents with Verbal Reinforcement Learning},
  author = {Noah Shinn and Federico Cassano and Edward Berman and Ashwin Gopinath and Karthik Narasimhan and Shunyu Yao},
  booktitle = {Advances in Neural Information Processing Systems},
  volume = {36},
  pages = {8634--8652},
  year = {2023},
  url = {https://proceedings.neurips.cc/paper_files/paper/2023/hash/1b44b878bb782e6954cd888628510e90-Abstract-Conference.html}
}

@inproceedings{zhao2023expel,
  title = {{ExpeL}: {LLM} Agents Are Experiential Learners},
  author = {Andrew Zhao and Daniel Huang and Quentin Xu and Matthieu Lin and Yong-Jin Liu and Gao Huang},
  booktitle = {Proceedings of the AAAI Conference on Artificial Intelligence},
  volume = {38},
  number = {17},
  pages = {19632--19642},
  year = {2024},
  doi = {10.1609/aaai.v38i17.29950},
  url = {https://ojs.aaai.org/index.php/AAAI/article/view/29950}
}

@article{wang2023voyager,
  title = {{Voyager}: An Open-Ended Embodied Agent with Large Language Models},
  author = {Guanzhi Wang and Yuqi Xie and Yunfan Jiang and Ajay Mandlekar and Chaowei Xiao and Yuke Zhu and Linxi Fan and Anima Anandkumar},
  journal = {Transactions on Machine Learning Research},
  year = {2024},
  url = {https://openreview.net/forum?id=ehfRiF0R3a}
}

@inproceedings{wang2025awm,
  title = {Agent Workflow Memory},
  author = {Zhiruo Wang and Jiayuan Mao and Daniel Fried and Graham Neubig},
  booktitle = {International Conference on Machine Learning},
  year = {2025},
  url = {https://openreview.net/forum?id=NTAhi2JEEE}
}

@misc{zhou2025memento,
  title = {Memento: Fine-tuning {LLM} Agents without Fine-tuning {LLM}s},
  author = {Huichi Zhou and Yihang Chen and Siyuan Guo and Xue Yan and Kin Hei Lee and Zihan Wang and Ka Yiu Lee and Guchun Zhang and Kun Shao and Linyi Yang and Jun Wang},
  year = {2025},
  eprint = {2508.16153},
  archivePrefix = {arXiv},
  primaryClass = {cs.LG},
  doi = {10.48550/arXiv.2508.16153},
  url = {https://arxiv.org/abs/2508.16153}
}

@misc{fang2025memp,
  title = {{MemP}: Exploring Agent Procedural Memory},
  author = {Runnan Fang and Yuan Liang and Xiaobin Wang and Jialong Wu and Shuofei Qiao and Pengjun Xie and Fei Huang and Huajun Chen and Ningyu Zhang},
  year = {2025},
  eprint = {2508.06433},
  archivePrefix = {arXiv},
  primaryClass = {cs.CL},
  doi = {10.48550/arXiv.2508.06433},
  url = {https://arxiv.org/abs/2508.06433}
}

@misc{zheng2025skillweaver,
  title = {{SkillWeaver}: Web Agents can Self-Improve by Discovering and Honing Skills},
  author = {Boyuan Zheng and Michael Y. Fatemi and Xiaolong Jin and Zora Zhiruo Wang and Apurva Gandhi and Yueqi Song and Yu Gu and Jayanth Srinivasa and Gaowen Liu and Graham Neubig and Yu Su},
  year = {2025},
  eprint = {2504.07079},
  archivePrefix = {arXiv},
  primaryClass = {cs.AI},
  doi = {10.48550/arXiv.2504.07079},
  url = {https://arxiv.org/abs/2504.07079}
}

@misc{wang2025sage,
  title = {Reinforcement Learning for Self-Improving Agent with Skill Library},
  author = {Jiongxiao Wang and Qiaojing Yan and Yawei Wang and Yijun Tian and Soumya Smruti Mishra and Zhichao Xu and Megha Gandhi and Panpan Xu and Lin Lee Cheong},
  year = {2025},
  eprint = {2512.17102},
  archivePrefix = {arXiv},
  primaryClass = {cs.AI},
  doi = {10.48550/arXiv.2512.17102},
  url = {https://arxiv.org/abs/2512.17102}
}

@misc{alzubi2026evoskill,
  title = {{EvoSkill}: Automated Skill Discovery for Multi-Agent Systems},
  author = {Salaheddin Alzubi and Noah Provenzano and Jaydon Bingham and Weiyuan Chen and Tu Vu},
  year = {2026},
  eprint = {2603.02766},
  archivePrefix = {arXiv},
  primaryClass = {cs.AI},
  doi = {10.48550/arXiv.2603.02766},
  url = {https://arxiv.org/abs/2603.02766}
}

@misc{tu2026d2skill,
  title = {Dynamic Dual-Granularity Skill Bank for Agentic {RL}},
  author = {Songjun Tu and Chengdong Xu and Qichao Zhang and Yaocheng Zhang and Xiangyuan Lan and Linjing Li and Dongbin Zhao},
  year = {2026},
  eprint = {2603.28716},
  archivePrefix = {arXiv},
  primaryClass = {cs.AI},
  doi = {10.48550/arXiv.2603.28716},
  url = {https://arxiv.org/abs/2603.28716}
}

@misc{li2026arise,
  title = {{ARISE}: Agent Reasoning with Intrinsic Skill Evolution in Hierarchical Reinforcement Learning},
  author = {Yu Li and Rui Miao and Zhengling Qi and Tian Lan},
  year = {2026},
  eprint = {2603.16060},
  archivePrefix = {arXiv},
  primaryClass = {cs.AI},
  doi = {10.48550/arXiv.2603.16060},
  url = {https://arxiv.org/abs/2603.16060}
}

@misc{liang2026skillnet,
  title = {{SkillNet}: Create, Evaluate, and Connect {AI} Skills},
  author = {Yuan Liang and Ruobin Zhong and Haoming Xu and Chen Jiang and Yi Zhong and Runnan Fang and Jia-Chen Gu and Shumin Deng and Yunzhi Yao and Mengru Wang and Shuofei Qiao and Xin Xu and Tongtong Wu and Kun Wang and Yang Liu and Zhen Bi and Jungang Lou and Yuchen Eleanor Jiang and Hangcheng Zhu and Gang Yu and Haiwen Hong and Longtao Huang and Hui Xue and Chenxi Wang and Yijun Wang and Zifei Shan and Xi Chen and Zhaopeng Tu and Feiyu Xiong and Xin Xie and Peng Zhang and Zhengke Gui and Lei Liang and Jun Zhou and Chiyu Wu and Jin Shang and Yu Gong and Junyu Lin and Changliang Xu and Hongjie Deng and Wen Zhang and Keyan Ding and Qiang Zhang and Fei Huang and Ningyu Zhang and Jeff Z. Pan and Guilin Qi and Haofen Wang and Huajun Chen},
  year = {2026},
  eprint = {2603.04448},
  archivePrefix = {arXiv},
  primaryClass = {cs.AI},
  doi = {10.48550/arXiv.2603.04448},
  url = {https://arxiv.org/abs/2603.04448}
}

@misc{ling2026agentskills,
  title = {Agent Skills: A Data-Driven Analysis of Claude Skills for Extending Large Language Model Functionality},
  author = {George Ling and Shanshan Zhong and Richard Huang},
  year = {2026},
  eprint = {2602.08004},
  archivePrefix = {arXiv},
  primaryClass = {cs.CL},
  doi = {10.48550/arXiv.2602.08004},
  url = {https://arxiv.org/abs/2602.08004}
}

@inproceedings{sentencebert2019,
  title = {{Sentence-BERT}: Sentence Embeddings using Siamese {BERT}-Networks},
  author = {Nils Reimers and Iryna Gurevych},
  booktitle = {Proceedings of the 2019 Conference on Empirical Methods in Natural Language Processing and the 9th International Joint Conference on Natural Language Processing},
  pages = {3982--3992},
  year = {2019},
  doi = {10.18653/v1/D19-1410},
  url = {https://aclanthology.org/D19-1410}
}

@misc{minisweagent2025,
  title = {{mini-SWE-agent}: A 100-line Software Engineering Agent},
  author = {{SWE-agent Team}},
  year = {2025},
  howpublished = {\url{https://github.com/SWE-agent/mini-swe-agent}},
  note = {Software repository}
}

@inproceedings{sweagent2024,
  title = {{SWE-agent}: Agent-Computer Interfaces Enable Automated Software Engineering},
  author = {John Yang and Carlos E. Jimenez and Alexander Wettig and Kilian Lieret and Shunyu Yao and Karthik Narasimhan and Ofir Press},
  booktitle = {Advances in Neural Information Processing Systems},
  year = {2024},
  url = {https://openreview.net/forum?id=5As8S9dyaD}
}

@article{deepseekmath2024grpo,
  title = {{DeepSeekMath}: Pushing the Limits of Mathematical Reasoning in Open Language Models},
  author = {Zhihong Shao and Peiyi Wang and Qihao Zhu and Runxin Xu and Junxiao Song and Mingchuan Zhang and Y. K. Li and Y. Wu and Daya Guo},
  journal = {arXiv preprint arXiv:2402.03300},
  year = {2024},
  url = {https://arxiv.org/abs/2402.03300}
}

@inproceedings{llmasjudge2023,
  title = {Judging {LLM}-as-a-Judge with {MT}-Bench and Chatbot Arena},
  author = {Lianmin Zheng and Wei-Lin Chiang and Ying Sheng and Siyuan Zhuang and Zhanghao Wu and Yonghao Zhuang and Zi Lin and Zhuohan Li and Dacheng Li and Eric P. Xing and Hao Zhang and Joseph E. Gonzalez and Ion Stoica},
  booktitle = {Advances in Neural Information Processing Systems},
  year = {2023},
  url = {https://arxiv.org/abs/2306.05685}
}

@inproceedings{lora2022,
  title = {{LoRA}: Low-Rank Adaptation of Large Language Models},
  author = {Edward J. Hu and Yelong Shen and Phillip Wallis and Zeyuan Allen-Zhu and Yuanzhi Li and Shean Wang and Lu Wang and Weizhu Chen},
  booktitle = {International Conference on Learning Representations},
  year = {2022},
  url = {https://openreview.net/forum?id=nZeVKeeFYf9}
}

@inproceedings{envbench2025,
  title = {{EnvBench}: A Benchmark for Automated Environment Setup},
  author = {Aleksandra Eliseeva and Alexander Kovrigin and Ilia Kholkin and Egor Bogomolov and Yaroslav Zharov},
  booktitle = {ICLR 2025 Workshop on Deep Learning for Code},
  year = {2025},
  url = {https://openreview.net/forum?id=izy1oaAOeX}
}

@inproceedings{swebench2024,
  title = {{SWE-bench}: Can Language Models Resolve Real-World {GitHub} Issues?},
  author = {Carlos E. Jimenez and John Yang and Alexander Wettig and Shunyu Yao and Kexin Pei and Ofir Press and Karthik Narasimhan},
  booktitle = {International Conference on Learning Representations},
  year = {2024},
  url = {https://openreview.net/forum?id=VTF8yNQM66}
}

@misc{swebenchverified2024,
  title = {Introducing {SWE-bench Verified}},
  author = {{OpenAI}},
  year = {2024},
  howpublished = {\url{https://openai.com/index/introducing-swe-bench-verified/}},
  note = {Accessed 2026-05-23}
}

@article{terminalbench2026,
  title = {{Terminal-Bench}: Benchmarking Agents on Hard, Realistic Tasks in Command Line Interfaces},
  author = {Mike A. Merrill and Alexander G. Shaw and Nicholas Carlini and Boxuan Li and Harsh Raj and Ivan Bercovich and Lin Shi and Jeong Yeon Shin and Thomas Walshe and E. Kelly Buchanan and Junhong Shen and Guanghao Ye and Haowei Lin and Jason Poulos and Maoyu Wang and Marianna Nezhurina and Jenia Jitsev and Di Lu and Orfeas Menis Mastromichalakis and Zhiwei Xu and Zizhao Chen and Yue Liu and others},
  journal = {arXiv preprint arXiv:2601.11868},
  year = {2026},
  url = {https://arxiv.org/abs/2601.11868}
}

@article{swesmith2025,
  title = {{SWE-smith}: Scaling Data for Software Engineering Agents},
  author = {John Yang and Kilian Lieret and Carlos E. Jimenez and Alexander Wettig and Kabir Khandpur and Yanzhe Zhang and Binyuan Hui and Ofir Press and Ludwig Schmidt and Diyi Yang},
  journal = {arXiv preprint arXiv:2504.21798},
  year = {2025},
  url = {https://arxiv.org/abs/2504.21798}
}

@inproceedings{react2023,
  title = {{ReAct}: Synergizing Reasoning and Acting in Language Models},
  author = {Shunyu Yao and Jeffrey Zhao and Dian Yu and Nan Du and Izhak Shafran and Karthik Narasimhan and Yuan Cao},
  booktitle = {International Conference on Learning Representations},
  year = {2023},
  url = {https://openreview.net/forum?id=WE_vluYUL-X}
}

@article{qwen3technicalreport2025,
  title = {{Qwen3} Technical Report},
  author = {{Qwen Team}},
  journal = {arXiv preprint arXiv:2505.09388},
  year = {2025},
  url = {https://arxiv.org/abs/2505.09388}
}

@misc{openai2026gpt54mini,
  title = {Introducing {GPT-5.4} mini and nano},
  author = {{OpenAI}},
  year = {2026},
  month = mar,
  howpublished = {\url{https://openai.com/index/introducing-gpt-5-4-mini-and-nano/}},
  note = {Accessed 2026-05-23}
}
\bibliographystyle{iclr2027_conference}

\newpage
\appendix

\section{Training Details}
\label{app:training}

This appendix provides additional details on the data construction and training configuration of \Name. We organize the discussion by training stage. Supervised fine-tuning first teaches the model the action schema and basic skill-management behavior, while reinforcement learning further optimizes the policy with downstream feedback. All training experiments are conducted on four H100 80GB GPUs.

\subsection{Supervised Fine-Tuning}
\label{app:sft}

\paragraph{Trajectory Sources.}
We construct SFT data from coding-agent trajectories collected across software-engineering tasks. The initial SWE-style trajectory pool is downloaded from publicly available mini-SWE-agent rollouts on SWE-smith, where Qwen3-30B is used as the coding agent. We additionally collect EnvBench trajectories using a ReAct-style agent. Since different sources use different logging formats, we normalize all trajectories into a unified representation containing task context, interleaved reasoning/action/observation steps, and final outcome information.

\begin{table}[h]
\centering
\small
\setlength{\tabcolsep}{6pt}
\caption{SFT data statistics by output action.}
\label{tab:app-sft-data}
\begin{tabular}{lr}
\toprule
Action & SFT examples \\
\midrule
task-level \texttt{generate} & 2419 \\
event-driven \texttt{generate} & 4485 \\
\texttt{skip} & 1344 \\
\texttt{evolve} & 1718 \\
\texttt{merge} & 1500 \\
\texttt{add} & 790 \\
\texttt{drop} & 600 \\
\midrule
Total & 12856 \\
\bottomrule
\end{tabular}
\end{table}

\begin{table}[h]
\centering
\small
\setlength{\tabcolsep}{6pt}
\caption{Training settings for \Name.}
\label{tab:app-training-settings}
\begin{tabular}{ll}
\toprule
Setting & Value \\
\midrule
\multicolumn{2}{l}{\textit{General settings}} \\
Backbone model & Qwen3.5-4B-Instruct \\
Training method & LoRA fine-tuning \\
Prompt template & \texttt{qwen3\_nothink} \\
Max sequence length & 14k tokens \\
Precision & bf16 \\
Hardware & 4$\times$H100 80GB \\
\midrule
\multicolumn{2}{l}{\textit{SFT settings}} \\
Epochs & 2 \\
batch size & 4 \\
Gradient accumulation & 8 \\
Learning rate & $1\times10^{-4}$ \\
Scheduler & cosine \\
Training time & $\sim$20 hours \\
\midrule
\multicolumn{2}{l}{\textit{RL settings}} \\
Group size & 6 generations per prompt \\
Gradient accumulation & 2 \\
Quality reward weight $\lambda$ & 0.25 \\
KL coefficient & 0.02 \\
Learning rate & $2\times10^{-6}$ \\
Training steps & 500 \\
Rollout temperature & 0.7 \\
Judge model & GPT-5.4-mini \\
Frozen coding policy & Qwen3.5-35B-A3B \\
Training time & $\sim$210 hours\\
\bottomrule
\end{tabular}
\end{table}

\begin{table*}[t]
\centering
\small
\setlength{\tabcolsep}{5pt}
\caption{RL training data statistics. Instances are counted within each phase, while the all-phases row reports global unique instance coverage. Prompt counts are grouped by action type.}
\label{tab:app-rl-data}
\begin{tabular}{lrrrrr}
\toprule
Phase & Instances & Task-level & Event-driven & Evolve & Maintain \\
\midrule
Phase 1 & 335 & 335 & 506 & 0 & 0 \\
Phase 2 & 513 & 199 & 302 & 487 & 0 \\
Phase 3 & 632 & 222 & 223 & 384 & 500 \\
\midrule
All phases & 660 & 756 & 1031 & 871 & 500 \\
\bottomrule
\end{tabular}
\end{table*}

\paragraph{Teacher Data Construction.}
We use teacher models to generate supervised targets for the skill-management action space. For task-level and event-driven extraction, the teacher reads normalized trajectory evidence and outputs either a reusable skill or a \texttt{skip} decision. For skill evolution, we synthesize training instances by retrieving an existing skill whose applicability condition or rules match a new trajectory, especially a failed or partially successful one. The paired trajectory is selected to expose missing conditions, better procedures, or failure modes not fully covered by the existing skill, and the teacher is asked to revise the skill or skip when no meaningful update is supported. This avoids collecting an additional round of skill-conditioned rollouts. For skill-bank maintenance, each candidate skill is paired with retrieved similar skills from the bank, and the teacher outputs \texttt{add}, \texttt{merge}, or \texttt{drop}.

\paragraph{SFT Configuration.}
We fine-tune Qwen3.5-4B-Instruct with LoRA using the \texttt{qwen3\_nothink} template\citep{lora2022}. The model is trained only on completion tokens. This stage teaches \Name to follow the action schema and produce valid extraction, evolution, and maintenance outputs before reinforcement learning. The main SFT hyperparameters are summarized in Table~\ref{tab:app-training-settings}.

\subsection{Reinforcement Learning}
\label{app:rl}

\paragraph{3-stage RL Training}
Since \Name learns multiple operation types and verifiable task feedback is sparse, we use a three-stage curriculum that follows the lifecycle of skill management and enables data reuse across stages. Phase 1 focuses on skill extraction from trajectory evidence. During this stage, generated skills are injected into the frozen coding policy to compute execution rewards, and the resulting skill-conditioned trajectories provide evidence for constructing phase-2 evolution prompts. Phase 2 trains extraction and evolution together, allowing \Name to revise existing skills with new trajectory evidence. Skills produced by the first two stages are then reused as candidate skills and bank entries for phase-3 maintenance prompts. Phase 3 contains the full action space, including extraction, evolution, and maintenance.

Table~\ref{tab:app-rl-data} summarizes the RL prompt data. Across all phases, the RL data covers 660 unique task instances. The curriculum starts with extraction-only prompts, then adds evolution prompts, and finally introduces maintenance prompts, aligning the training data with the skill-bank lifecycle.

\paragraph{RL Configuration.}
We optimize \Name with GRPO starting from the SFT checkpoint. The RL curriculum runs for 500 update steps in total, with 130, 120, and 250 steps for phases 1, 2, and 3, respectively. The frozen downstream coding policy used for reward-side rollouts is Qwen3.5-35B-A3B, and rubric-based quality and alignment judgments use GPT-5.4-mini. The main RL hyperparameters are summarized in Table~\ref{tab:app-training-settings}.

\section{Reward Design}
\label{appen:reward}
To compute the execution reward, we first pre-compute a no-skill baseline for every RL task instance. Specifically, we run the frozen Qwen3.5-35B-A3B coding policy four times on each instance without injected skills, and use the average verifier score as the baseline score. The execution reward of a generated skill is then computed as the improvement over this instance-specific baseline.

For rubric-based rewards, we design action-specific judge prompts for task-level extraction, event-driven extraction, skill evolution, skill-bank maintenance, and behavior alignment. Representative judge templates are provided in Figures~\ref{fig:task-level-judge}--\ref{fig:align-judge}. Each rubric consists of multiple dimensions, and each dimension is further decomposed into binary yes/no questions. The final rubric reward is computed as the number of satisfied yes/no questions divided by the total number of yes/no questions in the rubric. Thus, the relative weight of each dimension is determined by the number of questions assigned to it, while the final reward remains normalized to $[0,1]$.

\section{Experimental Details}
\label{app:exp-details}
\paragraph{Skill Retrieval.}
The frozen downstream coding policy solves each task with retrieved skills injected as prior knowledge. We build separate retrieval indexes for task-level and event-driven skills, so that the two granularities are matched with different query signals. For each skill, the retrieval document is constructed from its \texttt{title}, \texttt{when\_to\_apply}, and \texttt{rules}. We encode retrieval documents with \texttt{sentence-transformers/all-MiniLM-L6-v2} and build dense indexes separately for each benchmark and skill type \citep{sentencebert2019}. Retrieval is scoped to the same benchmark and same granularity, and skills generated from the same evaluation instance are filtered out to avoid same-instance leakage.

For task-level skills, the query is constructed from the task goal, problem statement, and available repository or benchmark context. Task-level skills are retrieved once before task solving and appended to the initial downstream policy user prompt as prior knowledge. For event-driven skills, the query is constructed online from local execution signals, including the current task context, recent reasoning, executed actions, observations, error messages, command outputs, and test-output snippets when available. Each instance is solved with retrieved skills through skill-conditioned generation, and every resulting skill-conditioned trajectory is prompted to \Name as evidence for skill evolution. Each newly extracted or evolved skill is then passed to the maintenance stage, where it may be added, merged, or dropped to update the current skill bank.

\paragraph{Baselines.}
We compare \Name with no-skill execution, prompt-based skill management, and subtask-level memory. The no-skill baseline uses the same frozen downstream coding policy without any prior-knowledge section. The prompt-based skill-management baseline is designed as an adaptation of recent automatic skill-bank methods to software-engineering tasks. Many such methods follow an extraction, evolution, and maintenance pipeline, or use a subset of these operations, but do not provide a software-engineering implementation. We therefore keep the same pipeline, action space, prompts, retrieval procedure, downstream policy, and decoding settings as \Name, while replacing the learned skill-management policy with fixed prompt-based decisions.

We also compare with a subtask-level memory baseline \citep{shen2026subtaskmemory}. This baseline is relevant to software-engineering tasks because long-horizon coding trajectories naturally decompose into smaller reasoning and action segments, and prior work has shown strong performance for reasoning-oriented memory in SWE settings. Since the original implementation is not publicly available, we implement a high-level reproduction. We segment each trajectory into subtasks using rule-based patterns based on agent action types and execution boundaries. Each sub-trajectory is then summarized into a memory item using a reasoning-oriented prompt, and relevant memory items are retrieved and injected into the downstream policy prompt in the same format used by other baselines.

\section{Skill Examples}
\label{app:skill-examples}

In general, an agent skill can be represented as a structured folder that contains reusable instructions, metadata, external resources, executable scripts, APIs, or tool-specific routines for guiding future agent behavior. In this work, since the coding-agent environments already provide fixed tools and execution interfaces, we follow prior work on instruction-level agent skills and focus on the natural-language instruction file rather than generating new scripts, APIs, or tools. Each skill file contains a short title, a granularity label, an applicability condition \texttt{when\_to\_apply}, and a set of actionable rules. \Name maintains two types of skills. Task-level skills encode high-level procedures for a task or a family of related tasks, while event-driven skills encode local guidance for recurring execution signals. Figures~\ref{fig:task-level-skill-example} and~\ref{fig:event-skill-example} show representative examples from the maintained skill bank.

\begin{figure*}[t]
\centering
\begin{skillbox}{Task-Level Skill Example}

\promptfield{Title}
Diagnose Maven build failures via environment, config, and dependency resolution.

\promptfield{Granularity}
\texttt{general}

\promptfield{When to Apply}
When a Java/Maven project fails to build and the cause may involve toolchain or wrapper configuration, repository settings, missing or stale dependencies, or build-phase ordering rather than source code errors.

\promptfield{Instructions}
\begin{itemize}[leftmargin=1.2em, itemsep=2pt, topsep=1pt]
    \item Start by identifying the project type and build instructions from repository metadata such as the main build file and README before making changes.
    \item Run the build with the project's Maven wrapper first, then capture both early and later output to distinguish dependency download progress from the actual failure.
    \item Use the first concrete failure message to distinguish network or repository issues, missing artifacts, and compilation errors before changing code.
    \item If the build fails in the enforcer or toolchain phase, use verbose output to identify the required Java and Maven versions before changing anything.
    \item Check the active Java and Maven toolchain versions in the shell and align them with the project's documented requirements when needed.
    \item When a dependency cannot be resolved, check whether the missing artifact is generated later in the build, and verify whether the build phase order or plugin lifecycle is appropriate.
    \item Inspect local Maven settings and repository configuration before changing code; misconfigured mirrors or missing settings files can mask the real build problem.
    \item After adjusting toolchain, wrapper settings, repository configuration, or dependency versions, rerun the build and re-check the logs to confirm whether the failure mode has changed.
\end{itemize}

\promptfield{Provenance}
This skill is an example from the maintained skill bank and was produced by a \texttt{merge} operation.

\end{skillbox}
\caption{Example task-level skill from the maintained skill bank.}
\label{fig:task-level-skill-example}
\end{figure*}

\begin{figure*}[t]
\centering
\begin{skillbox}{Event-Driven Skill Example}

\promptfield{Title}
Handle missing Maven test-jar dependencies.

\promptfield{Granularity}
\texttt{event-driven}

\promptfield{When to Apply}
When a Maven build fails with a dependency resolution error for a test-jar or similar scoped artifact that has not yet been produced or installed.

\promptfield{Instructions}
\begin{itemize}[leftmargin=1.2em, itemsep=2pt, topsep=1pt]
    \item Inspect the failing module's POM to confirm the dependency type and scope, for example a test-jar with compile scope.
    \item Check the producing module's POM to see whether it already defines an execution for the required artifact via a plugin configuration.
    \item If the test jar is only generated when tests run, either run the tests or explicitly invoke the Maven test-jar goal to create the test artifact before rebuilding.
    \item If the artifact is missing, add or configure the appropriate plugin execution in the producing module's POM to generate the required artifact, and place the new plugin configuration inside the correct parent element so the POM remains well formed.
    \item Do not simply retry the same Maven install command with tests skipped; skipping tests often prevents the test jar from being created.
    \item After generating or configuring the artifact, run a full install/build command so the artifact is generated and installed into the local repository; a compile-only run may not install the needed artifact.
    \item Verify that the artifact is actually present in the local Maven repository before retrying the original failing command.
    \item After the artifact is installed, rerun the original failing command to confirm that the dependency resolution error is resolved.
\end{itemize}

\promptfield{Provenance}
This skill is an example from the maintained skill bank and was produced by a \texttt{merge} operation.

\end{skillbox}
\caption{Example event-driven skill from the maintained skill bank.}
\label{fig:event-skill-example}
\end{figure*}

\section{Prompts}
We present the prompt template of task-level extraction in \ref{fig:task-level-prompt}, event-level extraction in \ref{fig:event-prompt}, skill evolution in \ref{fig:evolve-prompt}, and skill management in \ref{fig:maintain-prompt}.
\label{appen:prompt}
\begin{figure*}[t]
\centering
\begin{promptbox}{Task-Level Skill Extraction Prompt}

\promptfield{System Prompt}
You are an expert at extracting reusable memory from bash-agent trajectories.
Extract one reusable \texttt{general} skill from multiple related trajectories of a bash-based agent.
Use only the provided evidence.

\promptfield{Input}
The user prompt provides:
\begin{itemize}[leftmargin=1.2em, itemsep=1pt, topsep=1pt]
    \item light task context
    \item 2--3 related trajectories
    \item optional result summaries
\end{itemize}

\promptfield{Task}
Choose exactly one action:
\begin{itemize}[leftmargin=1.2em, itemsep=1pt, topsep=1pt]
    \item \texttt{generate}: create one bootstrap skill
    \item \texttt{skip}: create no skill
\end{itemize}

\promptfield{Rules}
\begin{enumerate}[leftmargin=1.4em, itemsep=2pt, topsep=1pt]
    \item A bootstrap skill is a task-level reusable pattern supported by multiple trajectories. It must be broader than a single local event.
    \item Generate only if the pattern is reusable across similar tasks. Keep \texttt{when\_to\_apply} high-level and write transferable, actionable \texttt{rules}.
    \item Ground every part of the skill in repeated evidence from the trajectories and outcomes. Do not invent unsupported steps, checks, or guidance. If failed trajectories reveal reusable cautions, include them as cautionary rules.
    \item Skip if the evidence is weak, contradictory, accidental, too local, or collapses into an event-level reaction instead of a task-level pattern.
    \item Do not include repository names, issue descriptions, exact task goals, variable names, function names, class names, module names, exact file paths, or one-off literals.
\end{enumerate}

\promptfield{Output Schema}
\texttt{generate}
\begin{lstlisting}[style=jsonstyle]
{
  "action": "generate",
  "skill": {
    "title": "short reusable skill name",
    "granularity": "general",
    "when_to_apply": "high-level task situation where this skill should be used",
    "rules": ["reusable rule 1", "reusable rule 2"]
  }
}
\end{lstlisting}

\texttt{skip}
\begin{lstlisting}[style=jsonstyle]
{
  "action": "skip",
  "reason": "short reason"
}
\end{lstlisting}

\end{promptbox}
\caption{Prompt template for task-level skill extraction.}
\label{fig:task-level-prompt}
\end{figure*}

\begin{figure*}[t]
\centering
\begin{promptbox}{Event-Driven Skill Extraction Prompt}

\promptfield{System Prompt}
You are an expert at extracting reusable memory from bash-agent trajectories.
Extract one reusable \texttt{event-driven} skill from a full trajectory of a bash-based agent.
Use only the provided evidence.

\promptfield{Input}
The user prompt provides:
\begin{itemize}[leftmargin=1.2em, itemsep=1pt, topsep=1pt]
    \item light task context
    \item one full trajectory
    \item optional result summary
\end{itemize}

\promptfield{Task}
Choose exactly one action:
\begin{itemize}[leftmargin=1.2em, itemsep=1pt, topsep=1pt]
    \item \texttt{generate}: create one event-driven skill
    \item \texttt{skip}: create no skill
\end{itemize}

\promptfield{Rules}
\begin{enumerate}[leftmargin=1.4em, itemsep=2pt, topsep=1pt]
    \item An event-driven skill is a local trigger-response pattern. It must focus on one important event inside the trajectory and stay narrower than a whole-task workflow.
    \item Generate only if the event yields a reusable lesson beyond this exact task. Keep \texttt{when\_to\_apply} transferable and write local actionable \texttt{rules}.
    \item Ground the trigger and guidance in the trajectory and result. Do not invent an event that is not clearly present. Failure-derived cautions are valid if they are clearly supported.
    \item If multiple candidate events exist, choose the single most reusable one.
    \item Skip if there is no strong reusable local event, or if the lesson is too task-specific or workflow-level.
    \item Do not include repository names, issue descriptions, exact task goals, variable names, function names, class names, module names, exact file paths, or one-off literals.
\end{enumerate}

\promptfield{Output Schema}
\texttt{generate}
\begin{lstlisting}[style=jsonstyle]
{
  "action": "generate",
  "skill": {
    "title": "short reusable skill name",
    "granularity": "event-driven",
    "when_to_apply": "high-level local signal or situation where this skill should be used",
    "rules": ["reusable rule 1", "reusable rule 2"]
  }
}
\end{lstlisting}

\texttt{skip}
\begin{lstlisting}[style=jsonstyle]
{
  "action": "skip",
  "reason": "short reason"
}
\end{lstlisting}

\end{promptbox}
\caption{Prompt template for event-driven skill extraction.}
\label{fig:event-prompt}
\end{figure*}

\begin{figure*}[t]
\centering
\begin{promptbox}{Skill Evolution Prompt}

\promptfield{System Prompt}
You are an expert at extracting and revising reusable memory from bash-agent trajectories.
You revise one existing reusable skill using new trajectory evidence from a bash-based agent.
Use only the provided evidence.

\promptfield{Input}
The user prompt provides:
\begin{itemize}[leftmargin=1.2em, itemsep=1pt, topsep=1pt]
    \item one or more relevant skills
    \item one trajectory
    \item optional result summary
\end{itemize}

\promptfield{Task}
Choose exactly one action:
\begin{itemize}[leftmargin=1.2em, itemsep=1pt, topsep=1pt]
    \item \texttt{evolve}: revise exactly one existing skill
    \item \texttt{skip}: make no revision
\end{itemize}

\promptfield{Rules}
\begin{enumerate}[leftmargin=1.4em, itemsep=2pt, topsep=1pt]
    \item \textbf{Alignment and need for revision.} Choose \texttt{evolve} only when one existing skill is clearly aligned with the current trajectory context and trigger pattern, and the current trajectory result shows that this aligned skill should be revised. If multiple skills are aligned and all are plausible revision targets, choose the single one that is most worth revising based on the strength and reusability of the evidence. The revision should be motivated by the observed outcome, such as a missing case, missing check, bad ordering, misleading guidance, or reusable caution.
    \item \textbf{Reusability and rule writing.} Revise the skill only if the trajectory reveals a reusable missing case, missing check, better decision rule, improved ordering, or reusable caution. \texttt{rules} should contain the revised reusable core of the skill as actionable guidance.
    \item \textbf{Applicability.} \texttt{when\_to\_apply} should remain at a high transferable level rather than concrete task text.
    \item \textbf{Grounding.} The revision must be directly supported by the provided trajectory and result. Do not add unsupported refinements.
    \item \textbf{Identity preservation.} Keep the same capability identity. Do not drift into a different skill.
    \item \textbf{When to skip.} Choose \texttt{skip} if the evidence is weak, contradictory, too local, or if the trajectory is better described as a brand-new skill rather than a revision.
    \item \textbf{Do not include task-specific details.} Do not include repository names, issue descriptions, exact task goals, variable names, function names, class names, module names, exact file paths, or one-off literals that only make sense for one instance.
\end{enumerate}

\promptfield{Output Schema}
\texttt{evolve}
\begin{lstlisting}[style=jsonstyle]
{
  "action": "evolve",
  "target_skill_id": "id of the single skill you chose to revise",
  "reason": "short reason",
  "skill": {
    "title": "short reusable skill name; same capability identity as target skill",
    "granularity": "general | event-driven",
    "when_to_apply": "high-level condition where this revised skill should be used",
    "rules": ["revised rule 1", "revised rule 2"]
  }
}
\end{lstlisting}

\texttt{skip}
\begin{lstlisting}[style=jsonstyle]
{
  "action": "skip",
  "reason": "short reason"
}
\end{lstlisting}

\end{promptbox}
\caption{Prompt Template for skill evolution.}
\label{fig:evolve-prompt}
\end{figure*}

\begin{figure*}[t]
\centering
\begin{promptbox}{Skill-Bank Maintenance Prompt}

\promptfield{System Prompt}
You are an expert at maintaining reusable memory for bash-based agents.
You decide how a candidate reusable skill should enter the skill bank for a bash-based agent.
Use only the provided candidate skill and retrieved similar skills.

\promptfield{Input}
The user prompt provides:
\begin{itemize}[leftmargin=1.2em, itemsep=1pt, topsep=1pt]
    \item 1 candidate skill
    \item multiple retrieved similar skills
\end{itemize}

\promptfield{Task}
Choose exactly one action:
\begin{itemize}[leftmargin=1.2em, itemsep=1pt, topsep=1pt]
    \item \texttt{add}: accept the candidate as a new skill
    \item \texttt{merge}: merge the candidate with exactly one retrieved skill
    \item \texttt{drop}: reject the candidate
\end{itemize}

\promptfield{Rules}
\begin{enumerate}[leftmargin=1.4em, itemsep=2pt, topsep=1pt]
    \item \textbf{Add.} Choose \texttt{add} when the candidate is reusable, coherent, and distinct from the retrieved skills.
    \item \textbf{Merge.} Choose \texttt{merge} when the candidate and one retrieved skill share the same capability identity and can be combined into one stronger reusable skill. The merged skill should have clearer applicability and cleaner rules with less duplication.
    \item \textbf{Drop.} Choose \texttt{drop} when the candidate is already covered by stronger retrieved skills, redundant, weakly evidenced, unsafe, too local, or too task-specific.
    \item \textbf{Do not include task-specific details.} No resulting skill may include repository names, issue descriptions, exact task goals, variable names, function names, class names, module names, exact file paths, or one-off literals that only make sense for one instance.
\end{enumerate}

\promptfield{Output Schema}
\texttt{add}
\begin{lstlisting}[style=jsonstyle]
{
  "action": "add",
  "reason": "short reason"
}
\end{lstlisting}

\texttt{drop}
\begin{lstlisting}[style=jsonstyle]
{
  "action": "drop",
  "reason": "short reason"
}
\end{lstlisting}

\texttt{merge}
\begin{lstlisting}[style=jsonstyle]
{
  "action": "merge",
  "merge_target_skill_id": "id of the retrieved skill selected for merge",
  "reason": "short reason",
  "skill": {
    "title": "short reusable skill name",
    "granularity": "general | event-driven",
    "when_to_apply": "high-level condition where this skill should be used",
    "rules": ["merged rule 1", "merged rule 2"]
  }
}
\end{lstlisting}

\noindent Schema requirements for \texttt{merge.skill}:
\begin{itemize}[leftmargin=1.2em, itemsep=1pt, topsep=1pt]
    \item \texttt{title}: short reusable name; not task-specific.
    \item \texttt{when\_to\_apply}: transferable applicability, not concrete task text.
    \item \texttt{rules}: a non-empty list of reusable actionable rules.
\end{itemize}

\end{promptbox}
\caption{Prompt template for skill-bank maintenance.}
\label{fig:maintain-prompt}
\end{figure*}

\begin{figure*}[t]
\centering
\begin{promptbox}{Task-Level Skill Quality Judge}
\promptfield{System Prompt}
You are an expert judge for skill-manager outputs. You are strict and critical by default. When in doubt, give a lower score.

\promptfield{Task}
Judge a proposed \texttt{bootstrap} prior-knowledge output. The output should be a task-level reusable item with \texttt{when\_to\_apply} and \texttt{rules}, rather than a local event-trigger rule, one-step trick, or instance-specific note.

\promptfield{Evaluation Input}
\texttt{\{\{TRAJECTORY\_EVIDENCE\}\}} and \texttt{\{\{PROPOSED\_OUTPUT\}\}}.

\promptfield{Scoring Rule}
For each dimension, answer the listed sub-questions with yes or no. The score is the number of yes answers.

\promptfield{Dimensions}
\begin{itemize}[leftmargin=1.2em, itemsep=2pt, topsep=1pt]
    \item \textbf{groundedness, 0--3.} Q1: every rule is directly traceable to observable trajectory behavior or outcome. Q2: \texttt{when\_to\_apply} is directly supported by trajectory signals. Q3: no rule is contradicted, absent, or weakly hinted.
    \item \textbf{reusability, 0--3.} Q1: rules avoid repository names, file paths, versions, or identifiers. Q2: rules transfer to at least two distinct project types. Q3: applicability is broad enough for unseen repositories but not nearly every coding task.
    \item \textbf{specificity, 0--3.} Q1: each rule gives a concrete executable action or check. Q2: rules add value beyond common engineering practice. Q3: rules would not equally apply to most unrelated coding tasks.
    \item \textbf{format\_validity, 0--1.} The output is valid and complete.
    \item \textbf{when\_to\_apply\_quality, 0--3.} Q1: applicability contains discriminating conditions. Q2: it would not trigger on clearly different tasks. Q3: it is narrow enough to avoid applying to most SWE tasks.
    \item \textbf{task\_level\_granularity, 0--3.} Q1: the output is a multi-step workflow-level pattern. Q2: it can guide a new agent from the start of a similar task. Q3: it is not primarily a local reactive rule.
\end{itemize}

\promptfield{Output Format}
\begin{lstlisting}[style=jsonstyle]
{
  "groundedness": 0,
  "reusability": 0,
  "specificity": 0,
  "format_validity": 0,
  "when_to_apply_quality": 0,
  "task_level_granularity": 0,
  "reason": "short explanation citing specific evidence or lack thereof"
}
\end{lstlisting}
\end{promptbox}
\caption{Rubric template for task-level skill quality judgment.}
\label{fig:task-level-judge}
\end{figure*}

\begin{figure*}[t]
\centering
\begin{promptbox}{Event-Driven Skill Quality Judge}
\promptfield{System Prompt}
You are an expert judge for skill-manager outputs. You are strict and critical by default. When in doubt, give a lower score.

\promptfield{Task}
Judge a proposed \texttt{event-driven} prior-knowledge output. The output should capture one reusable local event and its trigger-response knowledge, rather than a whole-task workflow or instance-specific note.

\promptfield{Evaluation Input}
\texttt{\{\{TRAJECTORY\_EVIDENCE\}\}} and \texttt{\{\{PROPOSED\_OUTPUT\}\}}.

\promptfield{Scoring Rule}
For each dimension, answer the listed sub-questions with yes or no. The score is the number of yes answers.

\promptfield{Dimensions}
\begin{itemize}[leftmargin=1.2em, itemsep=2pt, topsep=1pt]
    \item \textbf{groundedness, 0--3.} Q1: every rule is traceable to observable behavior or outcome. Q2: the trigger appears explicitly in the trajectory. Q3: no rule is contradicted, absent, or weakly hinted.
    \item \textbf{reusability, 0--3.} Q1: rules avoid repository names, file paths, exact error text, or other non-transferable identifiers. Q2: rules apply to the same event type in distinct projects. Q3: the trigger can recur in unseen projects.
    \item \textbf{specificity, 0--3.} Q1: each rule gives a concrete action or check. Q2: rules add event-specific value beyond generic debugging. Q3: rules would not apply to most unrelated local events.
    \item \textbf{format\_validity, 0--1.} The output is valid and complete.
    \item \textbf{when\_to\_apply\_quality, 0--3.} Q1: the trigger is specific and recognizable. Q2: it would not fire on unrelated local events. Q3: it is narrow enough to avoid frequent false triggers.
    \item \textbf{event\_level\_granularity, 0--3.} Q1: the output focuses on one local trigger-response pattern. Q2: it is local and reactive rather than workflow-level. Q3: the trigger is distinguishable from other common events.
\end{itemize}

\promptfield{Output Format}
\begin{lstlisting}[style=jsonstyle]
{
  "groundedness": 0,
  "reusability": 0,
  "specificity": 0,
  "format_validity": 0,
  "when_to_apply_quality": 0,
  "event_level_granularity": 0,
  "reason": "short explanation citing specific evidence or lack thereof"
}
\end{lstlisting}
\end{promptbox}
\caption{Rubric template for event-driven skill quality judgment.}
\label{fig:event-judge}
\end{figure*}

\begin{figure*}[t]
\centering
\begin{promptbox}{Skill Evolution Quality Judge}
\promptfield{System Prompt}
You are an expert judge for skill-manager outputs. You are strict and critical by default. When in doubt, give a lower score.

\promptfield{Task}
Judge a proposed \texttt{evolve} update. The output should refine existing prior knowledge using new trajectory evidence, rather than paraphrase the old skill, drift to a new skill, or produce an instance-specific note.

\promptfield{Evaluation Input}
\texttt{\{\{SYSTEM\_PROMPT\}\}}, \texttt{\{\{EXISTING\_PRIOR\_KNOWLEDGE\}\}}, \texttt{\{\{TRAJECTORY\_EVIDENCE\}\}}, and \texttt{\{\{PROPOSED\_OUTPUT\}\}}.

\promptfield{Scoring Rule}
For each dimension, answer the listed sub-questions with yes or no. The score is the number of yes answers.

\promptfield{Dimensions}
\begin{itemize}[leftmargin=1.2em, itemsep=2pt, topsep=1pt]
    \item \textbf{groundedness, 0--3.} Q1: every changed rule is supported by new evidence. Q2: changed applicability is supported without unjustified broadening. Q3: no new claim is contradicted, absent, or weakly hinted.
    \item \textbf{reusability, 0--3.} Q1: the update avoids one-off identifiers. Q2: changed guidance applies to distinct similar future situations. Q3: applicability lets a future agent decide when not to use the skill.
    \item \textbf{specificity, 0--3.} Q1: changed rules give concrete actions, checks, orderings, or decision criteria. Q2: the update adds value beyond universal debugging. Q3: the guidance is domain-specific enough that unrelated domains would not fit.
    \item \textbf{format\_validity, 0--1.} The output is valid and complete.
    \item \textbf{failure\_responsiveness, 0--3.} Q1: new evidence reveals a failure, limitation, contradiction, missing case, or caution. Q2: the update directly addresses it. Q3: the update would change a future concrete decision.
    \item \textbf{update\_quality, 0--3.} Q1: the update preserves skill identity. Q2: it adds, corrects, narrows, or sharpens non-redundant content. Q3: the final skill is coherent and focused.
\end{itemize}

\promptfield{Output Format}
\begin{lstlisting}[style=jsonstyle]
{
  "groundedness": 0,
  "reusability": 0,
  "specificity": 0,
  "format_validity": 0,
  "failure_responsiveness": 0,
  "update_quality": 0,
  "reason": "short explanation"
}
\end{lstlisting}
\end{promptbox}
\caption{Rubric template for skill evolution quality judgment.}
\label{fig:evolve-judge}
\end{figure*}

\begin{figure*}[t]
\centering
\begin{promptbox}{Skill-Bank Maintenance Merge Judge}

\promptfield{System Prompt}
You are an expert judge for skill-manager \texttt{maintain} outputs. You are strict and critical by default. When in doubt, give a lower score.

The action under review is \texttt{merge}. The model decided to merge the candidate prior knowledge into one specific existing entry, identified by \texttt{merge\_target\_skill\_id}, producing a single merged \texttt{skill} that replaces the target.

\noindent Note: maintain decisions are made without a trajectory. The input is the candidate prior-knowledge item and the existing prior-knowledge entries. Judge the merge purely on the textual evidence of the two sides and the merged result.

\promptfield{Task}
Judge a proposed \texttt{merge} decision in skill-bank maintenance. The goal is to merge a candidate with one substantially overlapping existing entry, producing a stronger entry that preserves both sides' useful content, avoids becoming an over-general umbrella, and remains high-quality reusable prior knowledge.

\promptfield{Evaluation Input}
\begin{itemize}[leftmargin=1.2em, itemsep=1pt, topsep=1pt]
    \item \texttt{\{\{CANDIDATE\_PRIOR\_KNOWLEDGE\}\}}
    \item \texttt{\{\{EXISTING\_PRIOR\_KNOWLEDGE\}\}}
    \item \texttt{\{\{PROPOSED\_OUTPUT\}\}}
\end{itemize}

\promptfield{Scoring Rule}
For each dimension, answer the listed sub-questions with yes or no. The score is the number of yes answers. A yes requires clear evidence. When in doubt, answer no.

\promptfield{Dimensions}
\begin{itemize}[leftmargin=1.2em, itemsep=2pt, topsep=1pt]
    \item \textbf{target\_choice\_correct, 0--3.} Q1: the chosen target's \texttt{when\_to\_apply} directly matches the candidate's trigger condition. Q2: the chosen target has the closest underlying capability among all existing entries. Q3: no other existing entry would be equally good or better.
    
    \item \textbf{target\_overlap\_substantive, 0--3.} Q1: candidate and target have at least 30\% conceptual rule overlap. Q2: their \texttt{when\_to\_apply} fields describe truly overlapping situations. Q3: keeping both separately would harm the bank through duplicate retrieval, conflict, or noise.
    
    \item \textbf{merge\_integration\_quality, 0--3.} Q1: overlapping rules are deduplicated and consolidated. Q2: the merged \texttt{when\_to\_apply} covers both original trigger conditions. Q3: the merged skill is internally consistent.
    
    \item \textbf{identity\_preserved, 0--3.} Q1: the merged title preserves the target's broad capability. Q2: the merged granularity matches or sensibly subsumes the target's granularity. Q3: the merged skill would be retrieved in roughly the same situations as the original target.
    
    \item \textbf{no\_critical\_loss\_target, 0--3.} Q1: every actionable rule from the original target is preserved, absorbed, or correctly deduplicated. Q2: target-specific orderings, exceptions, or cautions remain represented. Q3: target edge cases and trigger qualifiers remain intact.
    
    \item \textbf{no\_critical\_loss\_candidate, 0--3.} Q1: every actionable rule from the candidate is preserved, absorbed, or deduplicated. Q2: the candidate's distinctive insights are preserved or strengthened. Q3: candidate-specific trigger conditions are reflected in the merged result.
    
    \item \textbf{merged\_when\_to\_apply\_tightness, 0--3.} Q1: the merged applicability is no broader than the union of candidate and target triggers. Q2: it retains enough cues for a future agent to decide when not to retrieve it. Q3: it contains no over-generalized trigger such as ``any build issue'' or ``any failure with logs''.
    
    \item \textbf{format\_validity, 0--1.} The output is structurally valid for a \texttt{merge} decision, including a valid \texttt{merge\_target\_skill\_id}, a present merged \texttt{skill}, and required fields \texttt{title}, \texttt{granularity}, \texttt{when\_to\_apply}, and non-empty \texttt{rules}.
\end{itemize}

\promptfield{Output Format}
\begin{lstlisting}[style=jsonstyle]
{
  "target_choice_correct": 0,
  "target_overlap_substantive": 0,
  "merge_integration_quality": 0,
  "identity_preserved": 0,
  "no_critical_loss_target": 0,
  "no_critical_loss_candidate": 0,
  "merged_when_to_apply_tightness": 0,
  "format_validity": 0,
  "reason": "short explanation"
}
\end{lstlisting}

\end{promptbox}
\caption{Rubric template for judging merge decisions in skill-bank maintenance.}
\label{fig:maintain-merge-judge}
\end{figure*}

\begin{figure*}[t]
\centering
\begin{promptbox}{Behavior Alignment Judge}
\promptfield{System Prompt}
You are a strict judge evaluating whether an AI agent's behavior reflects provided prior knowledge. Be conservative when in doubt. Prior knowledge that only restates general best practices is hard to credit.

\promptfield{Task}
Judge whether a policy-model rollout actually reflects the provided prior knowledge. Alignment means the agent's concrete actions and reasoning show that the prior knowledge specifically guided behavior, not merely that the agent was competent.

\promptfield{Evaluation Input}
\texttt{\{\{TASK\_CONTEXT\}\}}, \texttt{\{\{USER\_PROMPT\}\}}, \texttt{\{\{PRIOR\_KNOWLEDGE\}\}}, \texttt{\{\{TRAJECTORY\_EVIDENCE\}\}}, and \texttt{\{\{RESULT\_SUMMARY\}\}}.

\promptfield{Scoring Dimensions}
\begin{itemize}[leftmargin=1.2em, itemsep=2pt, topsep=1pt]
    \item \textbf{when\_to\_apply\_match, 0--2.} A: the task context falls within the stated condition. B: concrete trajectory evidence shows the applicable situation is present.
    \item \textbf{rule\_specificity, 0--3.} A: a specific non-trivial rule maps to a concrete action. B: the action is plausibly caused by the prior knowledge rather than generic competence. C: the agent follows the intended spirit of the rule.
    \item \textbf{trajectory\_evidence, 0--3.} A: reasoning explicitly reflects the prior knowledge. B: at least two distinct steps align with distinct rules. C: alignment is sustained rather than coincidental.
\end{itemize}

\promptfield{Important Rules}
Do not reward task success, trajectory length, or apparent competence. Generic exploration does not earn credit unless the prior knowledge is specific and distinctive.

\promptfield{Output Format}
\begin{lstlisting}[style=jsonstyle]
{
  "when_to_apply_match": 0,
  "rule_specificity": 0,
  "trajectory_evidence": 0,
  "reason": "concise summary of the key factors driving your scores"
}
\end{lstlisting}
\end{promptbox}
\caption{Rubric template for behavior alignment judgment.}
\label{fig:align-judge}
\end{figure*}

\end{document}